%% file: main_boil_NeurIPS2020.tex
\documentclass[english,letterpaper]{article}
\pdfoutput=1
\usepackage[LGR,T1]{fontenc}
\usepackage[latin9]{inputenc}
\usepackage{babel}
\usepackage{float}
\usepackage{wrapfig}
\usepackage{booktabs}
\usepackage{amsmath}
\usepackage{amsthm}
\usepackage{amssymb}
\usepackage{graphicx}
\usepackage{xargs}[2008/03/08]
\usepackage[unicode=true,pdfusetitle,
 bookmarks=true,bookmarksnumbered=false,bookmarksopen=false,
 breaklinks=false,pdfborder={0 0 1},backref=false,colorlinks=false]
 {hyperref}

\makeatletter

\DeclareRobustCommand{\greektext}{%
  \fontencoding{LGR}\selectfont\def\encodingdefault{LGR}}
\DeclareRobustCommand{\textgreek}[1]{\leavevmode{\greektext #1}}
\ProvideTextCommand{\~}{LGR}[1]{\char126#1}

\providecommand{\tabularnewline}{\\}
\floatstyle{ruled}
\newfloat{algorithm}{tbp}{loa}
\providecommand{\algorithmname}{Algorithm}
\floatname{algorithm}{\protect\algorithmname}

\usepackage{algolyx}

\usepackage[nonatbib, final]{neurips_2020}
\usepackage{amsbsy}

\date{}

\usepackage{colortbl} 
\definecolor{header_color}{rgb}{1.0,0.93,0.87}
\definecolor{even_color}{rgb}{0.9,0.9,0.9}
\definecolor{subheader_color}{rgb}{0.85,0.93,0.95}
\definecolor{childheader_color}{rgb}{0.74,0.88,0.91}
\definecolor{cell_color}{rgb}{1.0,0.93,0.87}

\usepackage{mathptmx}
\usepackage{algolyx}

\usepackage{flushend}

\usepackage{colortbl} 
\definecolor{hd_cl}{rgb}{0.74,0.88,0.91}
\definecolor{header_color}{rgb}{0.74,0.88,0.91}
\definecolor{ev_cl}{rgb}{0.9,0.9,0.9}
\definecolor{even_color}{rgb}{0.9,0.9,0.9}
\definecolor{subhd_cl}{rgb}{0.85,0.93,0.95}
\definecolor{subheader_color}{rgb}{0.85,0.93,0.95}
\definecolor{childheader_color}{rgb}{1.0,0.93,0.87}

\makeatother

\input{macros.tex}

\title{Bayesian Optimization for Iterative Learning}

\newcommand*\samethanks[1][\value{footnote}]{\footnotemark[#1]}
\author{
Vu Nguyen \thanks{These authors contributed equally.}  \\ University of Oxford \\  vu@robots.ox.ac.uk
\And
Sebastian Schulze \samethanks \\ University of Oxford\\  sebastian.schulze@eng.ox.ac.uk
 \And
 Michael A. Osborne \\ University of Oxford \\  mosb@robots.ox.ac.uk
 }

\begin{document}

\maketitle

\begin{abstract}
The performance of deep (reinforcement) learning systems crucially
depends on the choice of hyperparameters. Their tuning is notoriously
expensive, typically requiring an iterative training process to run
for numerous steps to convergence. Traditional tuning algorithms only
consider the final performance of hyperparameters acquired after many
expensive iterations and ignore intermediate information from earlier
training steps. In this paper, we present a Bayesian optimization (BO)
approach which exploits the iterative structure of learning algorithms
for efficient hyperparameter tuning. We propose to learn an evaluation
function compressing learning progress at any stage of the training
process into a single numeric score according to both training success
and stability. Our BO framework is then balancing the benefit of assessing
a hyperparameter setting over additional training steps against their
computation cost. We further increase model efficiency by selectively
including scores from different training steps for any evaluated hyperparameter
set. We demonstrate the efficiency of our algorithm by tuning hyperparameters
for the training of deep reinforcement learning agents and convolutional
neural networks. Our algorithm outperforms all existing baselines
in identifying optimal hyperparameters in minimal time.

\end{abstract}

\section{Introduction}

\input{Introduction.tex}

\section{Related Work in Iteration-Efficient Bayesian Optimization\label{sec:preliminary}}

\input{RelatedWork.tex}

\section{Bayesian Optimization for Iterative Learning (BOIL)}

\input{Framework_Oct.tex}

\section{Experiments \label{sec:Experiment}}

\input{Experiments_v2.tex}

\section{Conclusion and Future work \label{sec:Conclusion-and-Future}}

\input{conclusion_v2.tex}

\section{Broader Impact}

Our work aims at making the optimization of processes operating in a step-wise fashion more efficient. As demonstrated this makes BOIL particularly well-suited to supporting supervised learning models and RL systems. By increasing training efficience of these models, we hope to contribute to their widespread deployment whilst reducing the computational and therefore environmental cost their implementation has.

Deep (reinforcement) learning systems find application in a wide range of settings that directly contribute to real world decisions, e.g., natural language processing, visual task, autonomous driving and many more. As machine learning models building on our contributions are being deployed in the real world, we encourage practicioners to put in place necessary supervision and override mechanisms as precautions against potential failure.

In a more general context, our algorithm may be seen as a step towards the construction of an automated pipeline for the training and deployment of machine learning models. A potential danger is that humans become further and further removed from the modelling process, making it harder to spot (potentially critical) failures. We do not see this as an argument against the construction of such a pipeline in principle, but instead encourage practicioners to reflect on potential biases indirectly encoded in the choice of data sets and models, they are feeding into said automated processes.

The growing opacity of machine learning models is a concern of its own and which automated training procedures will only contribute to. Opposing this is a rapidly growing corpus of work addressing the interpretability of trained machine learning models and their decision making. These can and should be used to rigorously analyse final training outcomes. Only then can we ensure that machine learning algorithm do indeed become a beneficial source of information guiding real world policy making as opposed to opaque, unquestioned entities.

While our main interest lies in the hyperparameter optimization of machine learning models, it should be noted that any iterative process depending on a set of parameters can make use of our contributions. Possible settings could, for instance, include the optimization of manufacturing pipelines in which factory setting are adjusted to increase productivity.

\section{Acknowledgements}
S. Schulze is supported by an I-CASE studentship funded by the EPSRC and Dyson.

\bibliographystyle{plain}
\bibliography{vunguyen}

\appendix
\input{NeurIPS2020_supplement.tex}
\end{document}

%% file: macros.tex
\global\long\def\bx{\mathbf{x}}%

\global\long\def\bz{\mathbf{z}}%

\global\long\def\bk{\mathbf{k}}%

\global\long\def\by{\mathbf{y}}%

\global\long\def\bK{\boldsymbol{K}}%

\global\long\def\and{\cap}%

\global\long\def\ess{\mathbb{E}}%

\newcommandx\ESS[2][usedefault, addprefix=\global, 1=]{\underset{#2}{\ess}\left[#1\right]}%

\global\long\def\idenmat{\mathbf{I}}%

%% file: Introduction.tex
Deep learning (DL) and deep reinforcement learning (DRL) have led
to impressive breakthroughs in a broad range of applications such
as game play \cite{mnih2013playing,silver2016mastering}, motor control
\cite{todorov2012mujoco}, and image recognition \cite{krizhevsky2012imagenet}.
To maintain general applicability, these algorithms expose sets of
hyperparameters to adapt their behavior to any particular task at
hand. This flexibility comes at the price of having to tune an additional
set of parameters -- poor settings lead to drastic performance losses
\cite{henderson2018deep,PB2,smith2018disciplined}. On top of being notoriously
sensitive to these choices, deep (reinforcement) learning systems
often have high training costs, in computational resources and time.
For example, a single training run on the Atari Breakout game took approximately
$75$ hours on a GPU cluster \cite{mnih2013playing}. Tuning DRL parameters
is further complicated as only noisy evaluations of an agent's final
performance are obtainable.

Bayesian optimization (BO) \cite{Hennig_2012Entropy,nguyen2019knowing, Shahriari_2016Taking} has recently achieved considerable success in optimizing these hyperparameters.
This approach casts the tuning process as a global optimization problem
based on noisy evaluations of a black-box function $f$. BO constructs
a surrogate model typically using a Gaussian process (GP) \cite{Rasmussen_2006gaussian},
over this unknown function.  This GP surrogate is used to build an
acquisition function \cite{Hernandez_2014Predictive,Wang_2017Max}
which suggests the next hyperparameter to evaluate. 

In modern machine learning (ML) algorithms \cite{Jordan_Mitchell_15machinelearning},
the training process is often conducted in an iterative manner.
A natural example is given by deep learning where training is usually
based on stochastic gradient descent and other iterative procedures.
Similarly, the training of reinforcement learning agents is mostly
carried out using multiple episodes. The knowledge accumulated during
these training iterations can be useful to inform BO. However, most
existing BO approaches \cite{Shahriari_2016Taking} define the objective
function as the average performance over the final training iterations.
In doing so, they ignore the information contained in the preceding
training steps.

In this paper, we present a Bayesian optimization approach for tuning
algorithms where iterative learning is available -- the cases of
deep learning and deep reinforcement learning. First, we consider
the joint space of input hyperparameters and number of training iterations
to capture the learning progress at different time steps in the training
process. We then propose to transform the whole training curve into
a numeric score according to user preference. To learn across the
joint space efficiently, we introduce a data augmentation technique
leveraging intermediate information from the iterative process. By
exploiting the iterative structure of training procedures, we encourage
our algorithm to consider running a larger number of cheap (but high-utility)
experiments, when cost-ignorant algorithms would only be able to run
a few expensive ones. We demonstrate the efficiency of our algorithm
on training DRL agents on several well-known benchmarks as well as
the training of convolutional neural networks. In particular, our
algorithm outperforms existing baselines in finding the best hyperparameter
in terms of wall-clock time. Our main contributions are:
\begin{itemize}
\item an algorithm to optimize the learning curve of a ML algorithm by using
training curve compression, instead of averaged final performance;
\item an approach to learn the compression curve from the data and a data
augmentation technique for increased sample-efficiency; 
\item demonstration on tuning DRL and convolutional neural networks. 
\end{itemize}

%% file: RelatedWork.tex
The first algorithm category employs stopping criteria to terminate some training
runs early and allocate resources towards more promising settings.
These criteria typically involve projecting towards a final score from early
training stages. Freeze-thaw BO \cite{swersky2014freeze} models the
training loss over time using a GP regressor under the assumption
that the training loss roughly follows an exponential decay. Based
on this projection, training resources are allocated to the most promising
settings. Hyperband \cite{falkner2018bohb,li2018hyperband} dynamically
allocates computational resources (e.g. training epochs or dataset
size) through random sampling and eliminates under-performing hyperparameter
settings by successive halving.

 Attempts have also been made to improve the epoch efficiency of
other hyperparameter optimization algorithms in \cite{dai2019bayesian,domhan2015speeding,klein2016learning} 
which predict the final learning outcome based on partially trained
learning curves to identify hyperparameter settings that are expected to under-perform and early-stop them. In the context of DRL, however, these stopping criteria, including the exponential decay assumed in Freeze-thaw BO \cite{swersky2014freeze}, may not be applicable, due to the unpredictable fluctuations of DRL reward curves. In the supplement, we illustrate the noisiness of DRL training.

The second category \cite{kandasamy2017multi,klein2017fast,li2018hyperband,Swersky_2013Multi,wu2019practical}
aims to reduce the resource consumption of BO by utilizing low-fidelity
functions which can be obtained by using a subset of the training
data or by training the ML model for a small number of iterations.
Multi-task BO \cite{Swersky_2013Multi} requires the user to define
a division of the dataset into pre-defined and discrete subtasks.
Multi-fidelity BO with continuous approximation (BOCA) \cite{kandasamy2017multi}
and hierarchical partition \cite{sen2018multi} extend this idea to
continuous settings. Specifically, BOCA first selects the hyperparameter
input and then the corresponding fidelity to be evaluated at. The
fidelity in this context refers to the use of different number of
learning iterations. Analogous to BOCA's consideration of continuous
fidelities, Fabolas \cite{klein2017fast} proposes to model the combined
space of input hyperparameter and dataset size and then select the optimal input and dataset size jointly.

The above approaches typically identify performance of hyperparameters
via the average (either training or validation) loss of the last learning
iterations. Thereby, they do not account for potential noise in the
learning process (e.g., they might select unstable settings that jumped
to high performance in the last couple of iterations). 

%% file: Framework_Oct.tex
\paragraph{Problem setting.}

We consider training a machine learning algorithm given a $d$-dimensional
hyperparameter $\bx\in\mathcal{X}\subset\mathcal{R}^{d}$ for $t$
iterations. This process has a training time cost $c(\bx,t)$ and
produces training evaluations $r(\cdot\mid\bx,t)$ for $t$ iterations, $t\in[T_{\min},T_{\max}]$.
 These could be episode rewards in DRL or training accuracies in DL.
An important property of iterative training is that we know the whole
curve at preceding steps $r(t' \mid \bx,t),$ $\forall t'\le t$.  

Given the raw training curve $r(\cdot\mid \bx,t)$, we assume an underlying
smoothed black-box function $f$, defined in Sec. \ref{subsec:Training-curve-compression}.
Formally, we aim to find $\bx^{*}=\arg\max_{\bx\in X}f(\bx,T_{\textrm{max}})$;
at the same time, we want to keep the overall training time, $\sum_{i=1}^{N}c(\bx_{i},t_{i})$,
of evaluated settings $[\bx_i, t_i]$ as low as possible. We summarize
our variables in Table $1$ in the supplement for ease of reading.

\subsection{Selecting a next point using iteration-efficient modeling \label{subsec:joint_modeling}}

We follow popular designs in \cite{klein2017fast,Krause_2011Contextual,song2019general,Swersky_2013Multi} and model the cost-sensitive black-box function as $f(\bx,t) \sim \text{GP}\left(0,k([\bx,t],[\bx',t'])\right)$, where $k$ is an appropriate covariance functions and $[\bx,t]\in\mathcal{R}^{d+1}$.
For simplicity and robustness, the cost function $c(\bx,t)$ is approximated by a linear regressor. Depending on the setting, it may be more appropriate to employ a second GP or different parametric model if the cost has a more complex dependence on hyperparameters $\bx$ and iterations $t$. We regularly (re-)optimize both kernel and cost function parameters in between point acquisitions.

More specifically, we choose the covariance function
as a product $k\left([\bx,t],[\bx',t']\right)=k(\bx,\bx')\times k(t,t')$
 to induce joint similarities over parameter and iteration space. 
We estimate the predictive mean and uncertainty for a GP \cite{Rasmussen_2006gaussian} at any
input $\bz_{*}=[\bx_{*},t_{*}]$ as\vspace{-15pt}

\begin{minipage}[t]{0.48\columnwidth}%
\begin{align}
\mu\left(\bz_{*}\right)= & \mathbf{k}_{*}\left[\mathbf{K}+\sigma_{y}^{2}\idenmat\right]^{-1}\mathbf{y}\label{eq:mu_f}
\end{align}
\end{minipage}%
\begin{minipage}[t]{0.48\columnwidth}%
\begin{align}
\sigma^{2}\left(\bz_{*}\right)= & k_{**}-\mathbf{k}_{*}\left[\mathbf{K}+\sigma_{y}^{2}\idenmat\right]^{-1}\mathbf{k}_{*}^{T}\label{eq:sigma_f}
\end{align}
\end{minipage}

where $\by=[y_{i}]_{\forall i}$,
$\ensuremath{\bk_{*}=[k\left(\bz_{*},\bz_{i}\right)]_{\forall i}}$,
$\ensuremath{\mathbf{K}=\left[k\left(\bz_{i},\bz_{j}\right)\right]_{\forall i,j}}$,
$\ensuremath{k_{**}=k\left(\bz_{*},\bz_{*}\right)}$, and $\sigma_{y}^{2}$ is the noise variance of $f$. Cost predictions at any particular parameter $\bx$ and time $t$ are given by $\mu_c([\bx_*,t_*])=\beta^T [\bx,t]$, where $\beta$ is directly computed from data $\{Z=[\bx_i,t_i], \mathbf{c}=[c_{i}]\}_{\forall i}$ as $\beta=(Z^T Z)^{-1}Z\mathbf{c}$ \cite{Bishop_2006pattern}.

\begin{figure}
    \includegraphics[width=0.72\columnwidth]{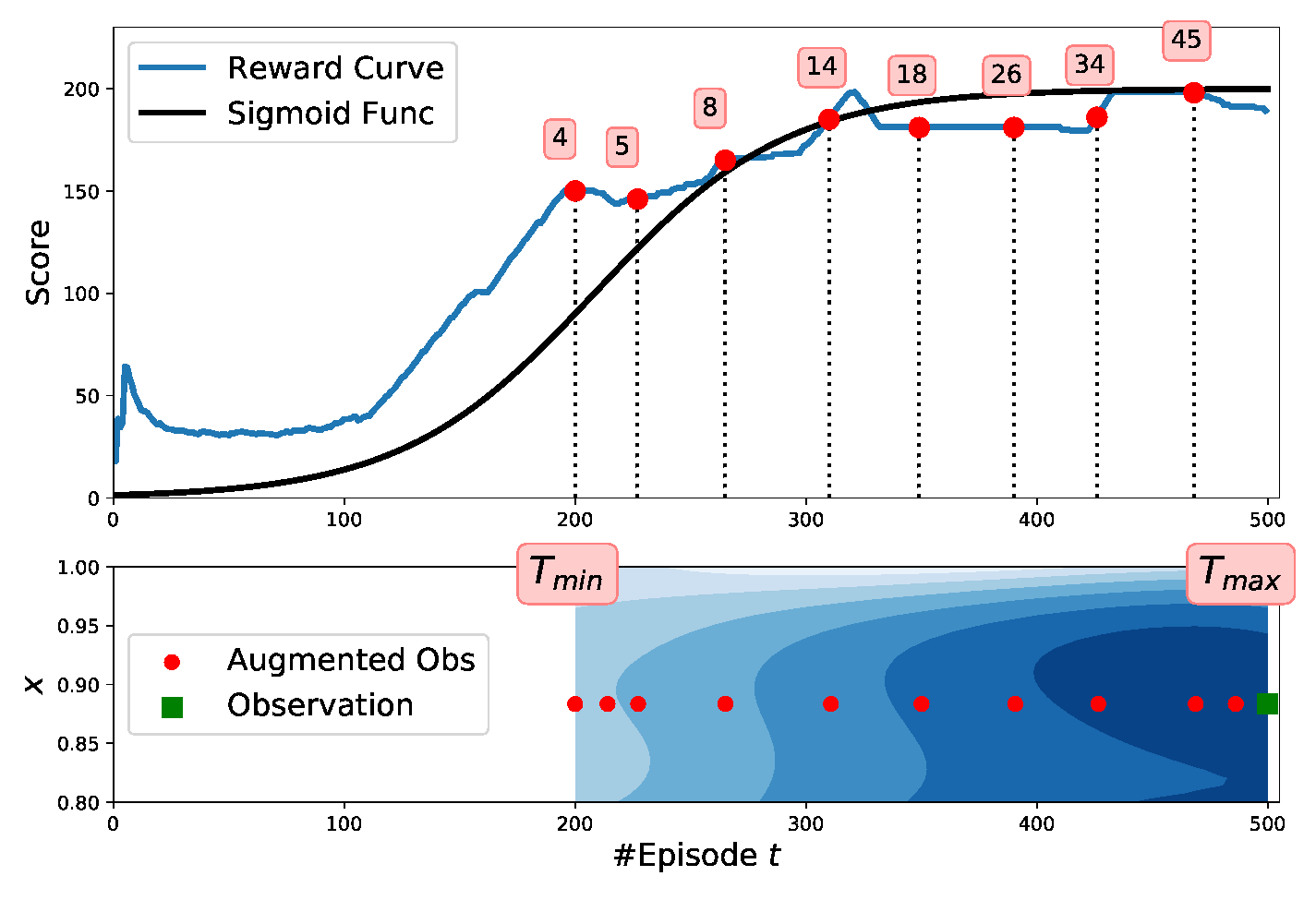}\includegraphics[width=0.27\textwidth]{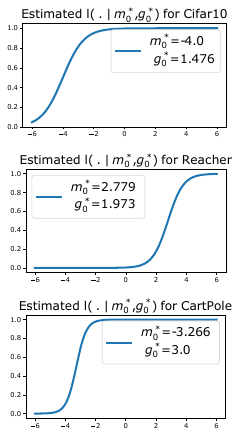}
\centering{}\caption{Left:  the score in pink box is a convolution of the reward curve $r(\cdot\mid \bx =0.9, t=500)$ and a Sigmoid function $l(u \mid g_0, m_0 )=\frac{1}{1+\exp\left(-g_{0}\left[u-m_{0}\right]\right)}$
up to time step t. Bottom: observations are selected to augment
the dataset (red dots). The heatmap indicates the GP predictive mean
$\mu$ for $f$ across the number of episodes $t$ used to train an agent.
$T_{\min}$ and $T_{\max}$ are two user-defined thresholds for the number
of training episodes. $x$ is a hyperparameter to be tuned.
Right: we learn the optimal parameter $g_{0}^{*}$ and $m_{0}^{*}$
for each experiment separately.\label{fig:VirtualObs_LogisticScore-3}}
\vspace{-10pt}
\end{figure}

Our goal is to select a point with high function value (exploitation),
high uncertainty (exploration) and low cost (cheap). At each iteration
$n$, we query the input parameter $\bx_{n}$ and the number of iteration
$t_{n}$ \cite{Snoek_2012Practical,wu2019practical}: 
\begin{align}
\bz_{n}=[\bx_{n},t_{n}] & =\underset{\bx\in\mathcal{X},t\in[T_{\min},T_{\max}]}{\arg\max}\alpha(\bx,t)/\mu_{c}(\bx,t).\label{eq:querying_new_point}
\end{align}
Although our framework is  available for any acquisition
choices \cite{Hernandez_2014Predictive,letham2019constrained,Wu_NIPS2016Parallel},
to cope with output noise, we follow \cite{Wang_2014Theoretical} and 
slight modify the expected improvement criterion using the maximum mean GP prediction $\mu_{n}^{\max}$
. Let $\lambda=\frac{\mu_{n}\left(\bz\right)-\mu_{n}^{\textrm{max}}}{\sigma_{n}\left(\bz\right)}$,
we then have a closed-form for the new expected improvement (EI) as $\alpha_{n}^{\textrm{EI}}\left(\bz\right)=\,\sigma_{n}\left(\bz\right)\phi\left(\lambda\right)+\left[\mu_{n}\left(\bz\right)-\mu_{n}^{\textrm{max}}\right]\Phi\left(\lambda\right)$
where $\phi$ is the standard normal p.d.f., $\Phi$ is the c.d.f,
$\mu_{n}$ and $\sigma_{n}$ are the GP predictive mean and variance
defined in Eq. (\ref{eq:mu_f}) and Eq. (\ref{eq:sigma_f}), respectively.

\subsection{Training curve compression and estimating the transformation function\label{subsec:Training-curve-compression}}

Existing BO approaches \cite{chen2018bayesian,li2018hyperband} typically
define the objective function as an average loss over the final learning
episodes. However, this does not take into consideration how stable
performance is or the training stage at which it has been achieved.
We argue that averaging learning losses is likely misleading due to
the noise and fluctuations of our observations (learning curves) --
particularly during the early stages of training. We propose to compress
the whole learning curve into a numeric score via a preference function
 representing the user's desired training curve. In the following,
we use the Sigmoid function (specifically the Logistic function) to
compute the utility score as
\begin{align}
y=\hat{y}\left(r,m_{0},g_{0}\right) & =r(\cdot \mid \bx,t)\bullet l(\cdot \mid m_{0},g_{0})=\sum_{u=1}^{t}\frac{r(u\mid \bx,t)}{1+\exp(-g_{0}\left[u-m_{0}\right])}\label{eq:marginalizing_curve}
\end{align}
where $\bullet$ is a dot product, a Logistic function $l(\cdot\mid m_{0},g_{0})$
is parameterized by a growth parameter $g_{0}$ defining a slope and
the middle point of the curve $m_{0}$. The optimal parameters $g_{0}$
and $m_{0}$ are estimated directly from the data. We illustrate
different shapes of $l$ parameterized by $g_{0}$ and $m_{0}$ in
the appendix. The Sigmoid preference has a  number of desirable properties.
As early weights are small, less credit is given to fluctuations
at the initial stages, making it less likely for our surrogate to
be biased towards randomly well performing settings. However, as weights
monotonically increase, hyperparameters with improving performance
are preferred. As weights saturate over time, stable, high performing
configurations are preferred over short ``performance spikes'' characteristic
of unstable training. Lastly, this utility score assigns higher values
to the same performance if it is being maintained over more episodes.

\paragraph{Learning the transformation function from data.}

Different compression curves $l()$, parameterized by different choices
of $g_{0}$ and $m_{0}$ in Eq. (\ref{eq:marginalizing_curve}), may
lead to different utilities $y$ and thus affect the performance.
The optimal values of $g_{0}^{*}$ and $m_{0}^{*}$ are unknown in
advance. Therefore, we propose to learn these values $g_{0}^{*}$
and $m_{0}^{*}$ directly from the data. Our intuition is that the
`optimal' compression curve $l(m_{0}^{*},g_{0}^{*})$ will lead to a
better fit of the GP. This better GP surrogate model, thus, will
result in better prediction as well as optimization performance.
We parameterize the GP log marginal likelihood $L$ \cite{Rasmussen_2006gaussian}
as the function of $m_{0}$ and $g_{0}$:
\begin{align}
L(m_{0},g_{0})= & \frac{1}{2}\hat{\by}^{T}\left(\bK+\sigma_{y}^{2}\idenmat\right)^{-1}\hat{\by}-\frac{1}{2}\ln\left|\bK+\sigma_{y}^{2}\idenmat\right|+\textrm{const}\label{eq:MarginalLLK}
\end{align}
where $\sigma_{y}^{2}$ is the output noise variance, $\hat{\by}$
is the function of $m_0$ and $g_{0}$ defined in Eq. (\ref{eq:marginalizing_curve}).
We optimize $m_{0}$ and $g_{0}$ (jointly with other GP hyperparameters)
using multi-start gradient descent. We derive the derivative $\frac{\partial L}{\partial m_{0}}=\frac{\partial L}{\partial\hat{y}}\frac{\partial\hat{y}}{\partial m_{0}}$
and $\frac{\partial L}{\partial g_{0}}=\frac{\partial L}{\partial\hat{y}}\frac{\partial\hat{y}}{\partial g_{0}}$
which can be computed analytically as:\vspace{-15pt}

\begin{center}
\begin{minipage}[t]{0.29\columnwidth}%
\begin{align*}
\frac{\partial L}{\partial\hat{y}} & =\left(\bK+\sigma_{y}^{2}\idenmat_{N}\right)^{-1}\hat{y};
\end{align*}
\end{minipage}\hfill{}%
\begin{minipage}[t]{0.35\columnwidth}%
\begin{align*}
\frac{\partial\hat{y}}{\partial m_{0}} & =\frac{-g_{0}\times\exp(-g_{0}\left[u-m_{0}\right])}{\left[1+\exp(-g_{0}\left[u-m_{0}\right])\right]^{2}};
\end{align*}
\end{minipage}\hfill{}%
\begin{minipage}[t]{0.35\columnwidth}%
\begin{align*}
\frac{\partial\hat{y}}{\partial g_{0}} & =\frac{-m_{0}\times\exp(-g_{0}\left[u-m_{0}\right])}{\left[1+\exp(-g_{0}\left[u-m_{0}\right])\right]^{2}}.
\end{align*}
\end{minipage}
\par\end{center}

The estimated compression curves are illustrated in Right Fig. \ref{fig:VirtualObs_LogisticScore-3}
and in Sec. \ref{subsec:Model--illustration}.

\subsection{Augmenting the training data\label{subsec:Augmentation-with-intermediate}}

When evaluating a parameter $\bx$ over $t$ iterations, we obtain
not only a final score but also all reward sequences $r(t' \mid \bx,t),\forall t'=1,...,t$.
The auxiliary information from the curve can be useful 
for BO. Therefore, we propose to augment the information from the
curve into the sample set of our GP model. 

A na\"ive approach for augmentation is to add a full curve of points
$\{[\bx,j],y_{j}\}_{j=1}^{t}$ where $y_{j}$ is computed  using
Eq. (\ref{eq:marginalizing_curve}). However, this approach can be redundant and may impose
serious issues in the conditioning of the GP covariance matrix. As
we cluster more evaluations closely, the conditioning of the GP covariance
degrades further, as discussed in \cite{mcleod2018optimization}.
This conditioning issue is especially serious in our noisy DRL settings.
\begin{figure}
\begin{centering}
\vspace{-0pt}
\par\end{centering}
\begin{centering}
\par\end{centering}
\begin{centering}
\includegraphics[width=0.5\columnwidth]{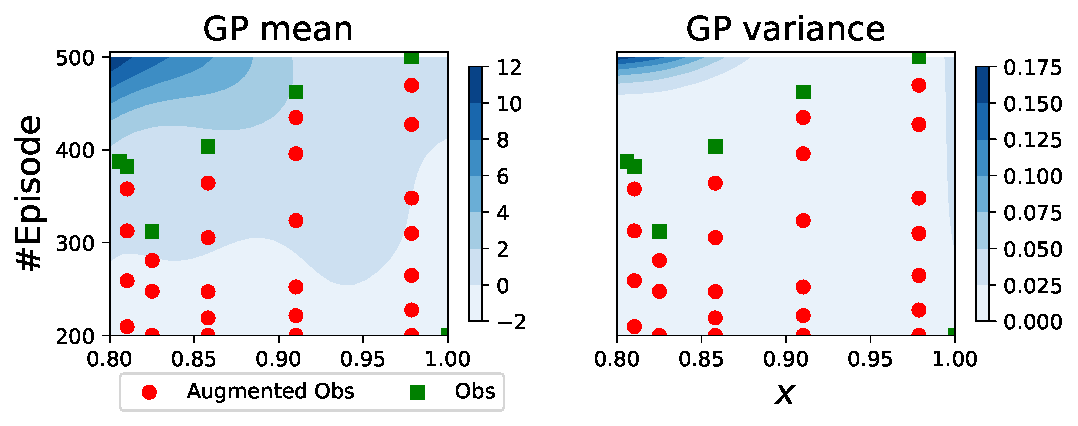}\includegraphics[width=0.5\columnwidth]{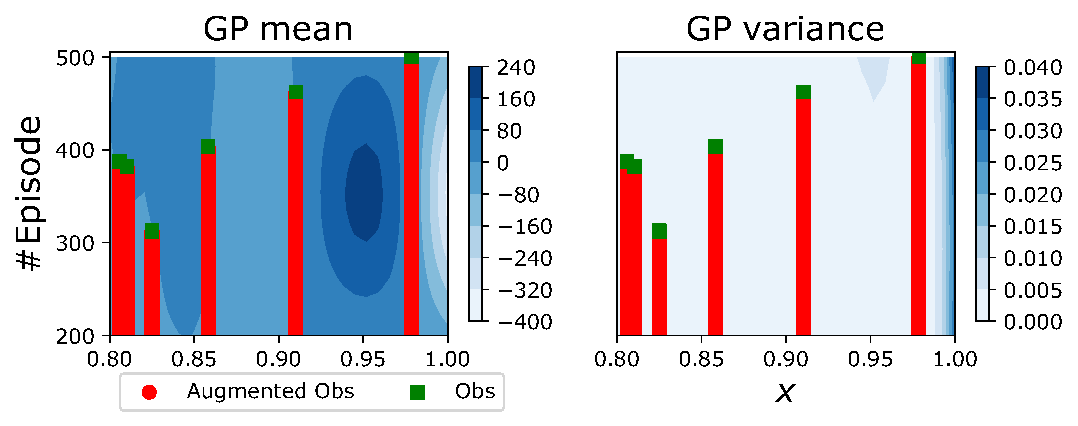}
\par\end{centering}
\begin{centering}
\vspace{-5pt}
\caption{GP with different settings. Left: our augmentation. Right: using
a full curve.  If we add too many observations, the GP covariance
matrix becomes ill-conditioned. On the right, the GP fit is poor with a large mean estimate range of $\left[-400,240\right]$
even though the output is standardized $\mathcal{N}(0,1)$. All x-axis
are over $x$, a hyperparameter to be tuned. \label{fig:GP_Example}}
\par\end{centering}
\vspace{-10pt}
\end{figure}
\begin{wrapfigure}{o}{0.47\columnwidth}%
\begin{centering}
\vspace{-5pt}
\includegraphics[width=0.47\columnwidth]{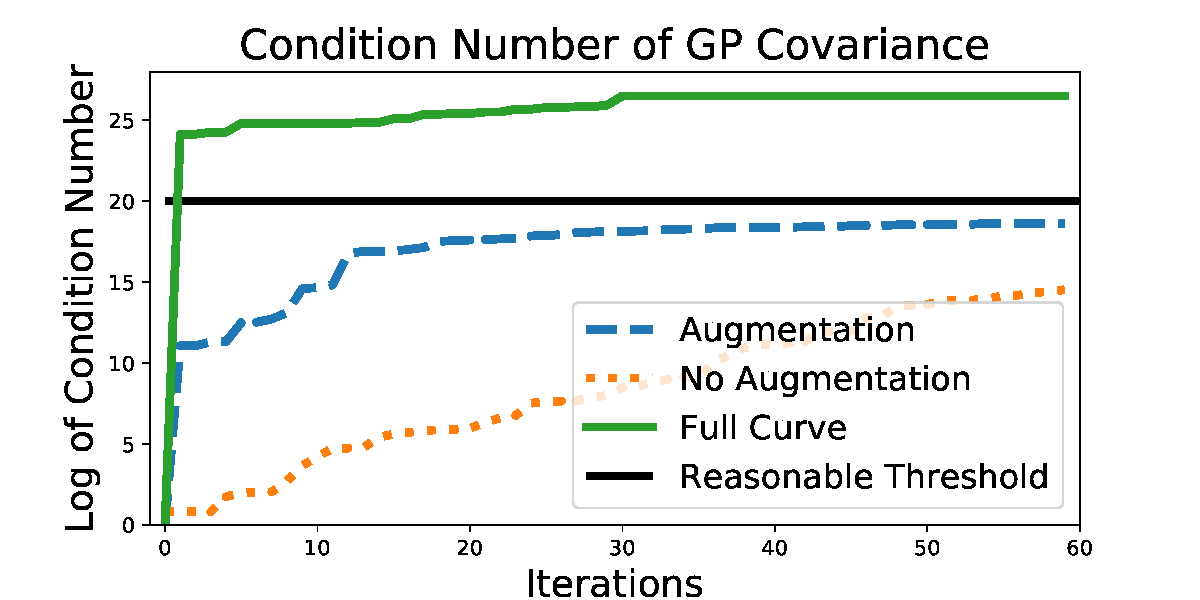}
\par\end{centering}
\begin{centering}
\vspace{-3pt}
\caption{The condition number of GP covariance matrix deteriorates if we add
the whole curve of points into a GP. The large condition number indicates
the nearness to singularity. \label{fig:Condition-number-of-1}}
\par\end{centering}
\vspace{-15pt}
\end{wrapfigure}We highlight this effect on GP estimation in Fig. \ref{fig:GP_Example}
wherein the GP mean varies erratically when the \emph{natural log} of the condition number of the GP covariance matrix goes above $25$ (see Fig. \ref{fig:Condition-number-of-1}) as we include the whole curve. 

\paragraph{Selecting subset of points from the curve.}
 Different solutions, such as the addition of artificial noise or altering the kernel's length-scales, have been proposed. We decide to use an active learning approach \cite{gal2017deep,osborne2012active} as sampled data points are expected to contain a lot of redundant information. As a consequence, the loss of information from sub-sampling the data should be minimal and information-eroding modification of the kernel matrix itself can be avoided. As a side benefit, the reduced number of sampled points speeds up inference in our GP models.


In particular, we select 
samples at the maximum of the GP predictive uncertainty.  Formally,
we sequentially select a set $Z=[z_{1},...z_{M}]$, $z_{m}=\left[\bx,t_{m}\right]$,
by varying $t_{m}$ while keeping $\bx$ fixed as
\begin{align}
z_{m}= & \arg\max_{\forall t'\le t}\thinspace\sigma([\bx,t']\mid D'),\forall m\le M\,\,\,\,\textrm{s.t}.\,\,\,\,\textrm{ln}\,\textrm{of\,cond}(K)\le\delta\label{eq:Augmentation}
\end{align}
where $D'=D\cup\{z_{j}=[\bx,t_{j}]\}_{j=1}^{m-1}$. This sub-optimisation
problem is done in a one-dimensional space of $t'\in\{T_{\textrm{min}},...,t\}$,
thus it is \emph{cheap} to optimize using (multi-start) gradient descent (the
derivative of GP predictive variance is available \cite{Rasmussen_2006gaussian}). Alternatively, a fixed-size grid could be considered, but this could cause conditioning issues when a point in the grid $\left[\bx,t_{\textrm{grid}}\right]$ is placed near
another existing point $\left[\bx',t_{\textrm{grid}}\right]$, i.e.,
$||\bx-\bx'||_{2}\le\epsilon$ for some small $\epsilon$.

These generated points $Z$ are used to calculate the output $r(z_{m})$
and augmented into the observation set $\left(X,Y\right)$ for fitting
the GP. The number of samples $M$ is adaptively chosen such
that the natural log of the condition number of the covariance matrix
is less than a threshold. This is  to ensure that the GP covariance
matrix condition number behaves well by reducing the number of unnecessary
points added to the GP at later stages.  We compute the utility score
$y_{m}$ given $z_{m}$ for each augmented point using Eq. (\ref{eq:marginalizing_curve}).
In addition, we can estimate the running time $c_{m}$ using the predictive
mean $\mu_{c}(z_{m})$.  We illustrate the augmented observations
and estimated scores in Fig. \ref{fig:VirtualObs_LogisticScore-3}.


\begin{algorithm}
\caption{Bayesian Optimization with Iterative Learning (BOIL)\label{alg:BO_DRL}}

\begin{algor}
\item [{{*}}] \textbf{Input}: \#iter $N$, initial data $D_{0}$, $\bz=\left[\bx,t\right]$.
\textbf{Output}: optimal $\bx^{*}$ and $y^{*}=\max_{\forall y\in D_{N}}y$
\end{algor}
\begin{algor}[1]
\item [{for}] $n=1....N$
\item [{{*}}] Fit a GP to estimate $\mu_{f}(),\sigma_{f}()$ 
from Eqs. (\ref{eq:mu_f},\ref{eq:sigma_f}) and a LR for cost $\mu_{c}()$
\item [{{*}}] Select $\bz_{n}=\arg\max_{\bx,t}\alpha(\bx,t)/\mu_{c}(\bx,t)$
and observe a curve $r$ and a cost $c$ from $f(\bz_{n})$
\item [{{*}}] Compressing the learning curve $r(\bz_{n})$ into numeric
score using Eq. (\ref{eq:marginalizing_curve}).
\item [{{*}}] Sample augmented points $\bz_{n,m},y_{n,m},c_{n,m},\forall m\le M$
given the curve and $D_{n}$ in Eq. (\ref{eq:Augmentation})
\item [{{*}}] Augment the data into $D_{n}$ and estimate Logistic curve
hyperparameters $m_{0}$ and $g_{0}$.
\item [{endfor}]~
\end{algor}
\end{algorithm}

 We summarize the overall algorithm in Alg. \ref{alg:BO_DRL}. To enforce non-negativity and numerical stability, we make use of the transformations $\alpha\leftarrow\log\left[1+\exp(\alpha)\right]$
and $\mu_{c}\leftarrow\log\left[1+\exp(\mu_{c})\right]$.

%% file: Experiments_v2.tex
We assess our  model by tuning hyperparameters for two DRL
agents on three environments and a CNN on two datasets. We provide
additional illustrations and experiments in the appendix.

\paragraph{Experimental setup.}

All experimental results are averaged over $20$ independent runs
with different random seeds.  Final performance is estimated by 
evaluating the chosen hyperparameter over the maximum number of iterations.
 All experiments are executed on a NVIDIA 1080 GTX GPU using the
tensorflow-gpu Python package. The DRL environments are available
through the OpenAI gym \cite{brockman2016openai} and Mujoco \cite{todorov2012mujoco}.
Our DRL implementations are based on the open source from Open AI
Baselines \cite{dhariwal2017openai}. We release our implementation at \url{https://github.com/ntienvu/BOIL}.

We use square-exponential kernels for the GP in our model and estimate
their parameters by maximizing the marginal likelihood \cite{Rasmussen_2006gaussian}.
We set the maximum number of augmented points to be $M=15$ and
a threshold for a natural log of GP condition number $\delta=20$.
We note that the optimization overhead is much less than the black-box
function evaluation time. 

\paragraph{Baselines.}

We compare with Hyperband \cite{li2018hyperband} which demonstrated
empirical successes in tuning deep learning applications in an iteration-efficient
manner. We extend the discrete multi-task BO \cite{Swersky_2013Multi} to the continuous case -- which can also be seen as continuous
multi-fidelity BO \cite{kandasamy2017multi,song2019general} as
in our setting, they both consider cost-sensitivity and iteration-efficiency. We, therefore, label the two baselines as continuous multi-task/fidelity
BO (CM-T/F-BO). We have ignored the minor difference in these settings,
such as multi-task approaches jointly optimizes the fidelity and input
while BOCA \cite{kandasamy2017multi} first selects the input and
then the fidelity. 

Our focus is to demonstrate the effectiveness of optimizing the learning curve using compression and augmentation techniques. We therefore omit the comparison of various acquisition functions and
kernel choices which can easily be used in our model. We also do not compare with Fabolas \cite{klein2017fast} which
is designed to vary dataset sizes, not iteration numbers. We would
expect the performance of Fabolas to be close to CM-T/F-BO. We are
unable to compare with FreezeThaw as the code is not available. However, the curves in our setting are not exponential decays and thus ill-suited to their model  (see last figure in the appendix). We have considered
an ablation study in the appendix using a time kernel following the exponential decay proposed in Freeze-thaw method \cite{swersky2014freeze}.

\paragraph{Task descriptions.}
We consider three DRL settings including a Dueling DQN (DDQN) \cite{wang2016dueling}
agent in the CartPole-v0 environment and Advantage Actor Critic (A2C)
\cite{mnih2016asynchronous} agents in the InvertedPendulum-v2 and
Reacher-v2 environments.  In addition to the DRL applications, we
tune $6$ hyperparameters for training a convolutional neural network
\cite{lecun1998gradient} on the SVHN dataset  and CIFAR10. Due
to space considerations, we  refer to the appendix for further 
details.

\subsection{Model  illustration\label{subsec:Model--illustration}}

\begin{wrapfigure}{o}{0.46\columnwidth}%
\begin{centering}
\vspace{-25pt}
\includegraphics[width=0.46\columnwidth]{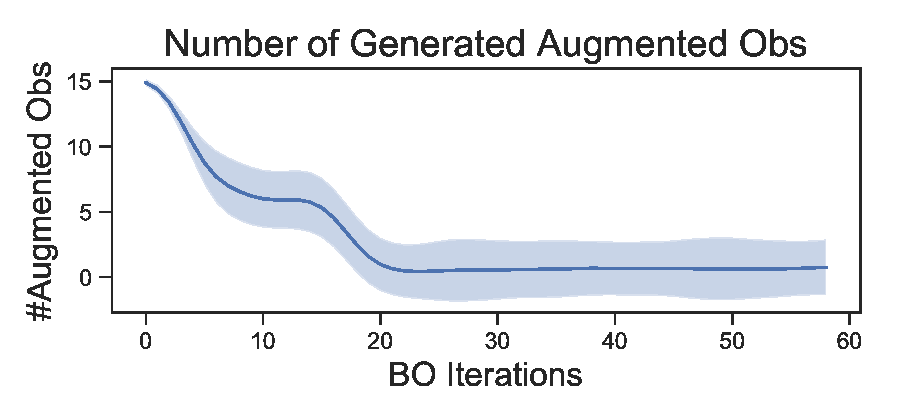}\caption{DDQN on CartPole. The number of augmented observations reduces
over time.\label{fig:CountVirtualObs}}
\par\end{centering}
\vspace{-10pt}
\end{wrapfigure}  We first illustrate the estimated compression function $l(m_{0}^{*},g_{0}^{*})$
in Right Fig. \ref{fig:VirtualObs_LogisticScore-3} from different
experiments. These Logistic parameters $g_{0}^{*}$ and $m_{0}^{*}$
are estimated by maximizing the GP marginal likelihood and used for
compressing the curve. We show that the estimated curve from CartPole
tends to reach the highest performance much earlier than Reacher because
CartPole is somewhat easier to train than Reacher.

We next examine the count of augmented observations generated per
iteration in Fig. \ref{fig:CountVirtualObs}. Although this number
is fluctuating, it tends to reduce over time. BOIL does not add more
augmented observations at the later stage when we have gained sufficient
information and GP covariance conditioning falls below our threshold
$\delta=20$.

\subsection{Ablation study of curve compression}

\begin{figure*}
\begin{centering}
\includegraphics[width=0.5\columnwidth]{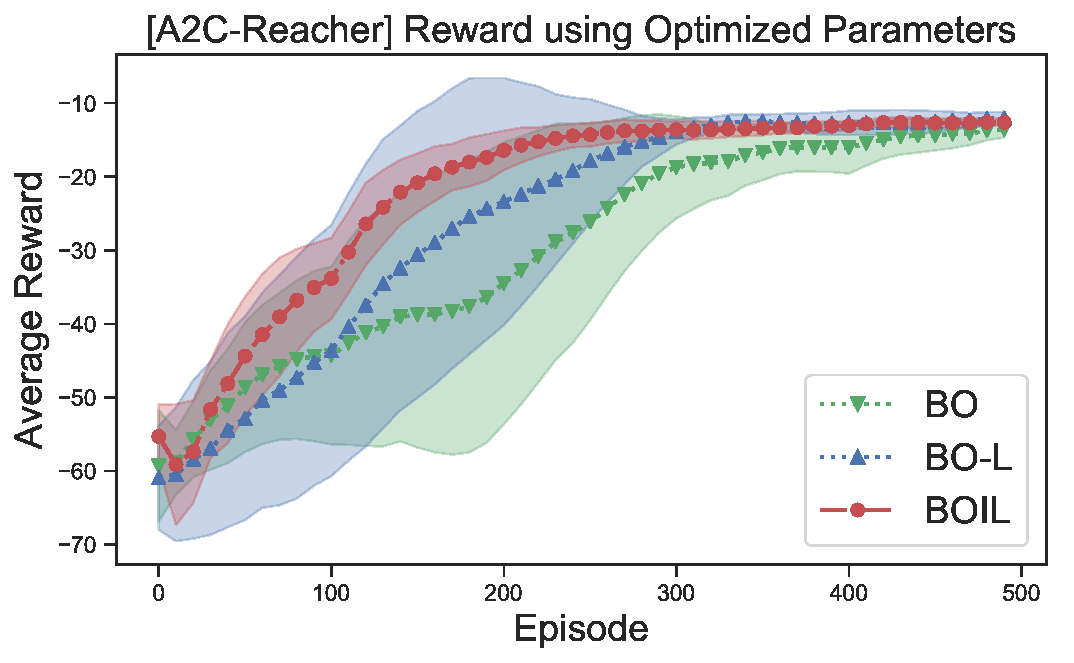}\hfill{}\includegraphics[width=0.5\columnwidth]{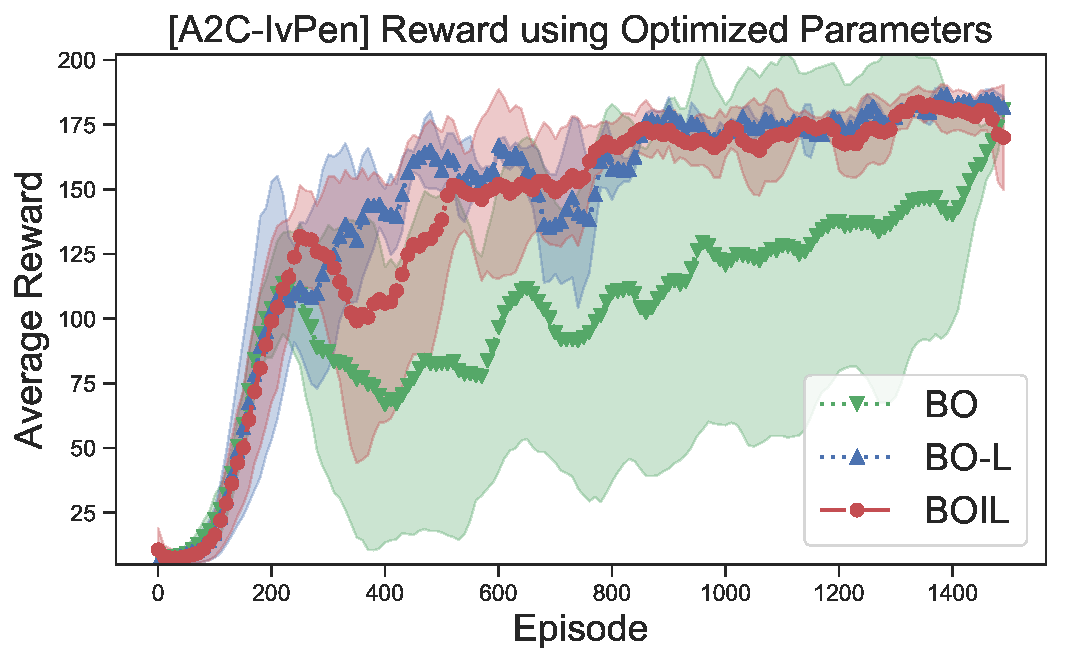}
\par\end{centering}
\caption{The learning curves of the best found parameters  by different approaches.
The curves show that BO-L and BOIL reliably identify parameters leading
to stable training. BOIL takes only half total time to find this optimal
curve. \label{fig:Reward-curves-using_best_parameters}}
\end{figure*}

To demonstrate the impact of our training curve compression, we compare
BOIL to vanilla Bayesian optimization (BO) and with compression (BO-L)
given the same number of iterations at $T_{\textrm{max}}$. We show
that using the curve compression leads to stable performance,
as opposed to the existing technique of averaging the last iterations.
We plot the learning curves of the best hyperparameters identified
by BO, BO-L and BOIL. Fig. \ref{fig:Reward-curves-using_best_parameters}
shows the learning progress over $T_{\max}$ episodes for each of
these. The curves are smoothed by averaging over $100$ consecutive
episodes for increased clarity. We first note that all three algorithms
eventually obtain similar performance at the end of learning. However,
since BO-L and BOIL take into account the preceding learning steps,
they achieve higher performance more quickly. Furthermore, they achieve
this more reliably as evidenced by the smaller error bars (shaded
regions).

\subsection{Tuning deep reinforcement learning and CNN}

We now optimize hyperparameters for deep reinforcement learning algorithms;
in fact, this application motivated the development of BOIL. The combinations
of hyperparameters to be tuned, target DRL algorithm and environment
can be found in the appendix.

\begin{figure*}
\begin{centering}
\includegraphics[width=0.5\columnwidth]{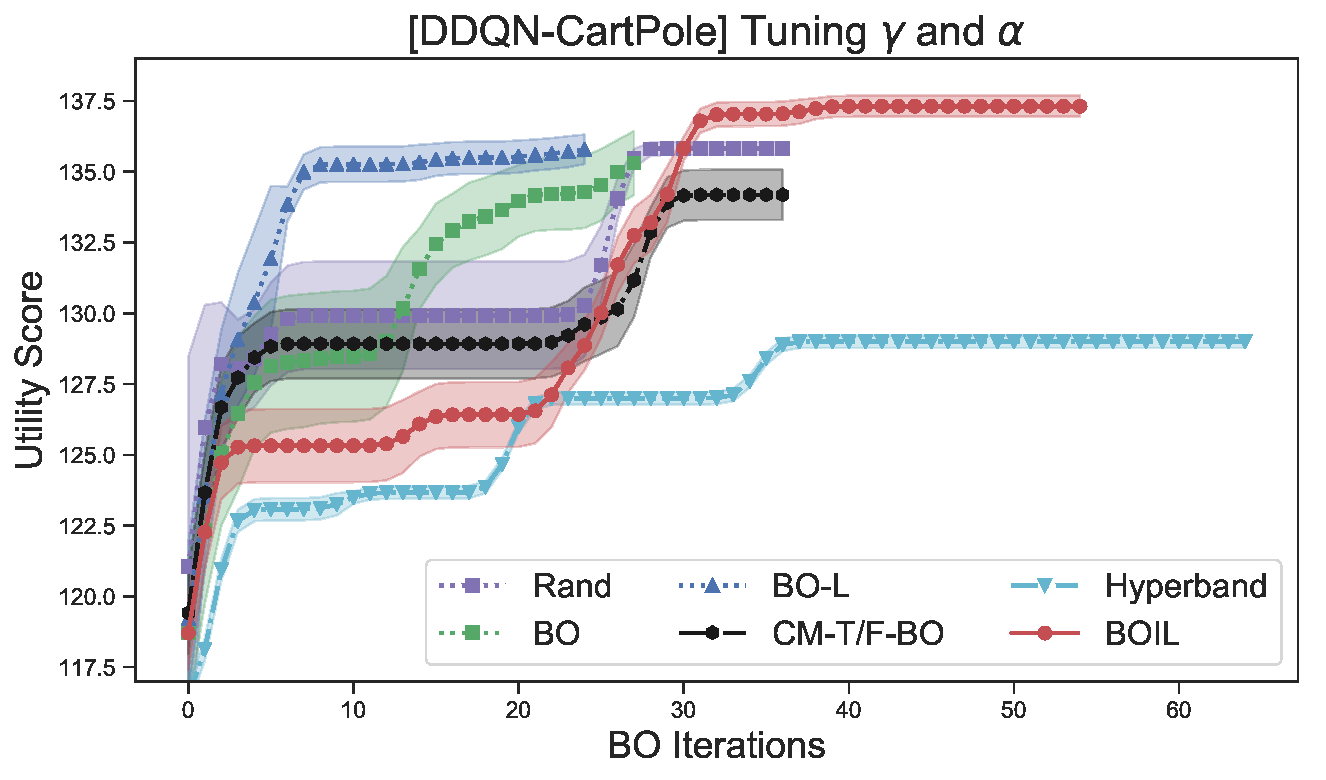}\includegraphics[width=0.5\columnwidth]{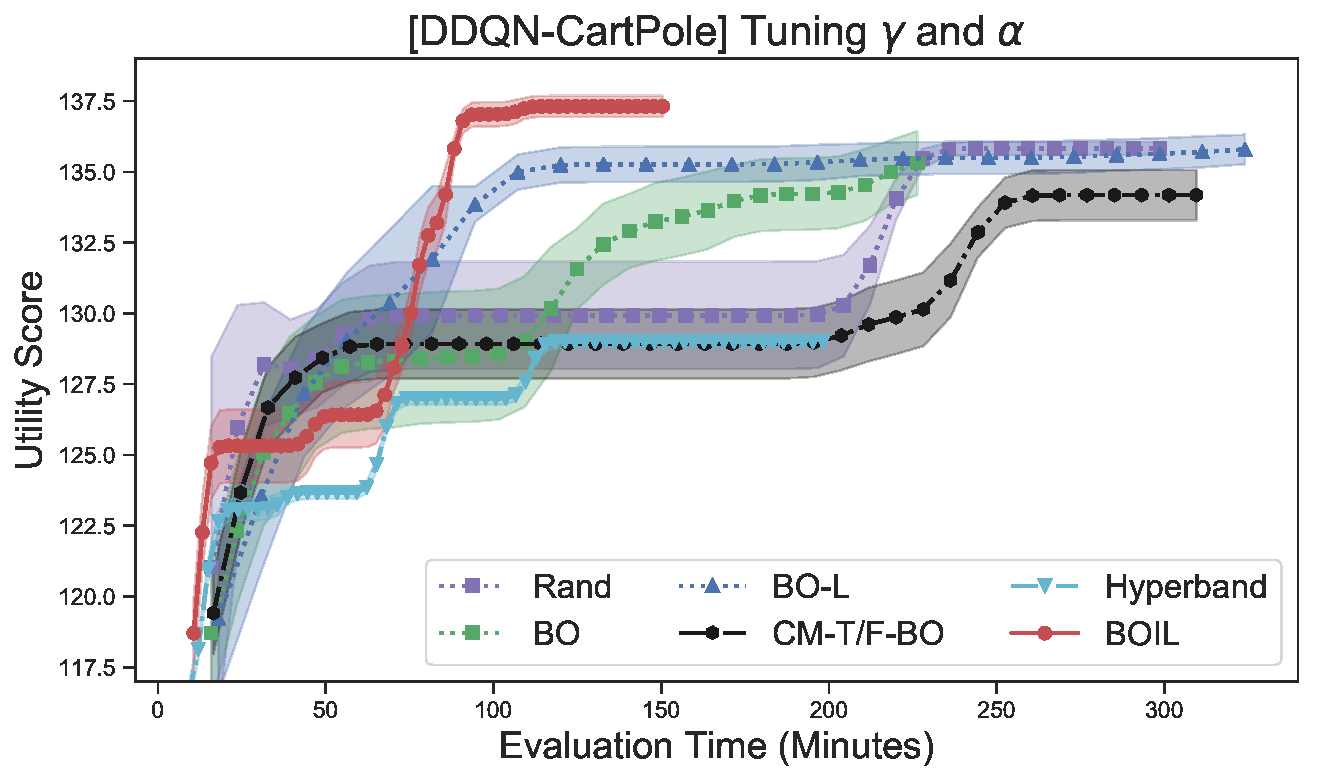}
\par\end{centering}
\begin{centering}
\includegraphics[width=0.5\columnwidth]{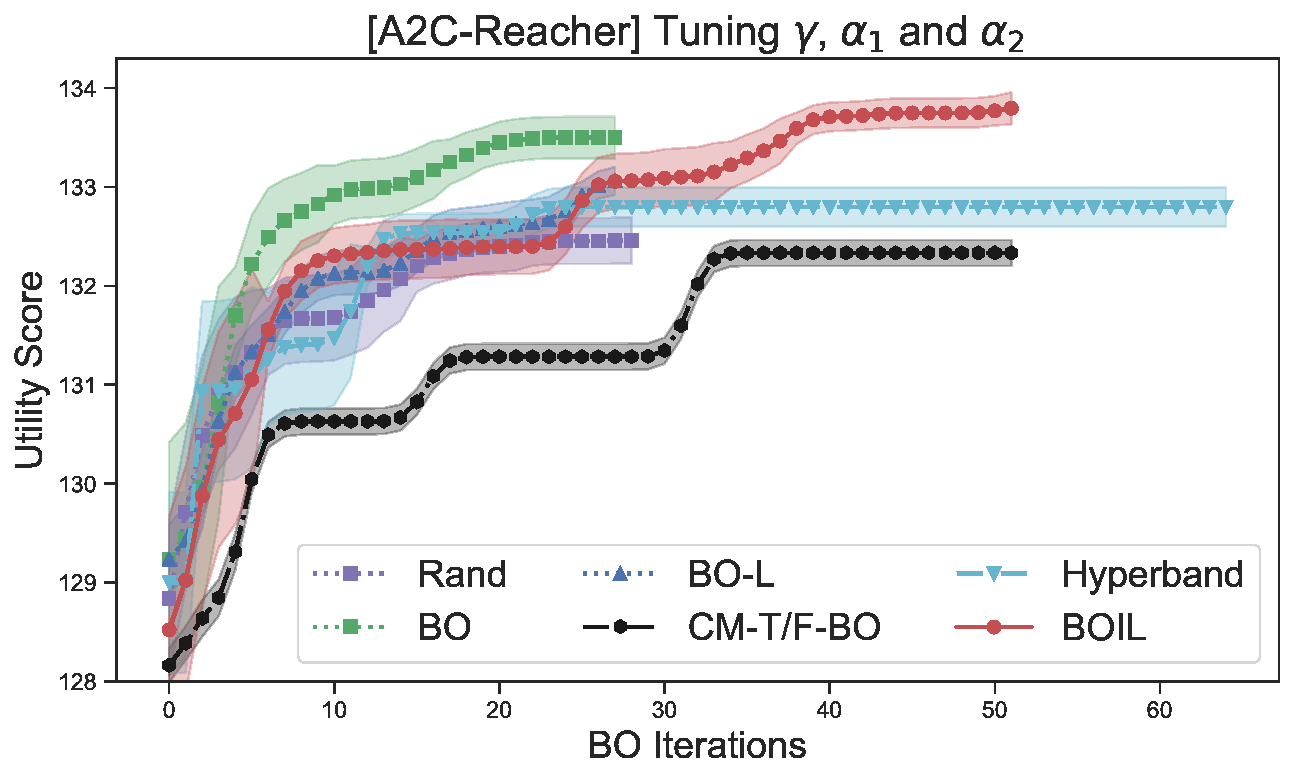}\includegraphics[width=0.5\columnwidth]{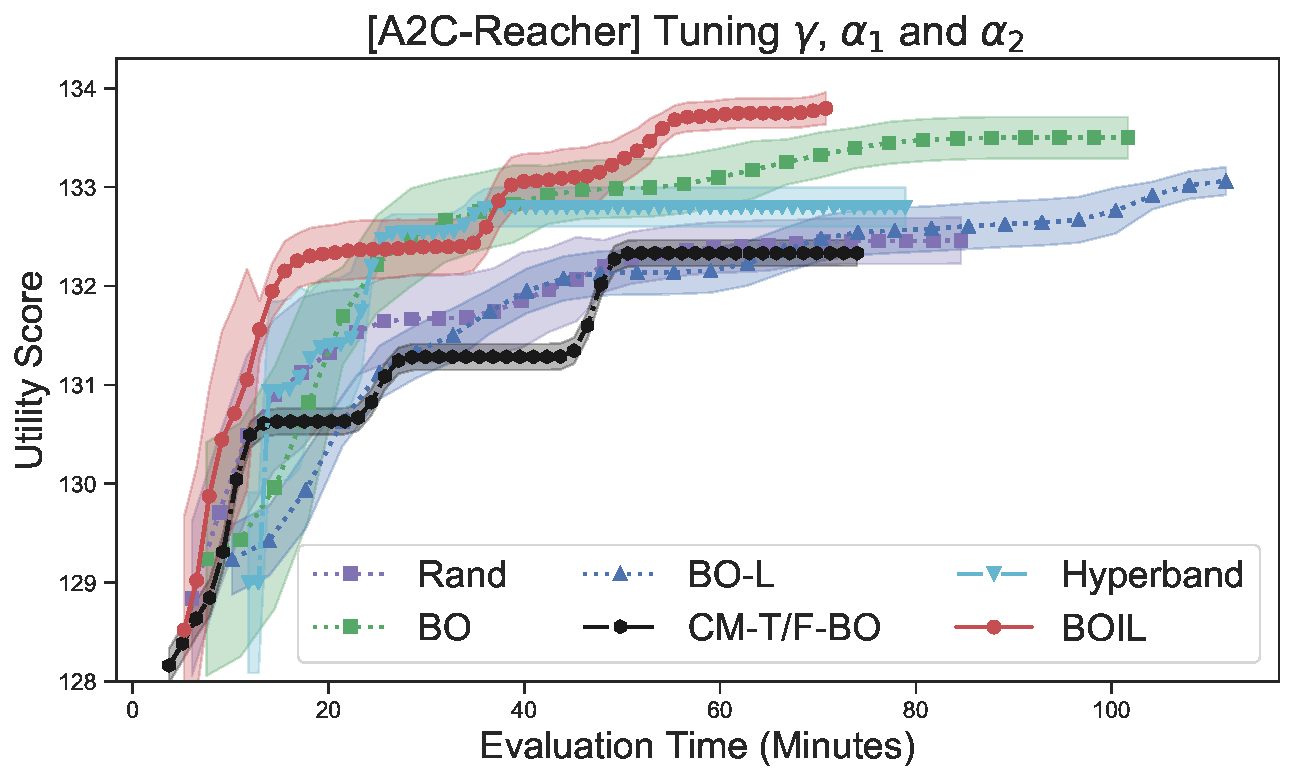}
\par\end{centering}
\begin{centering}
\par\end{centering}
\begin{centering}
\includegraphics[width=0.5\columnwidth]{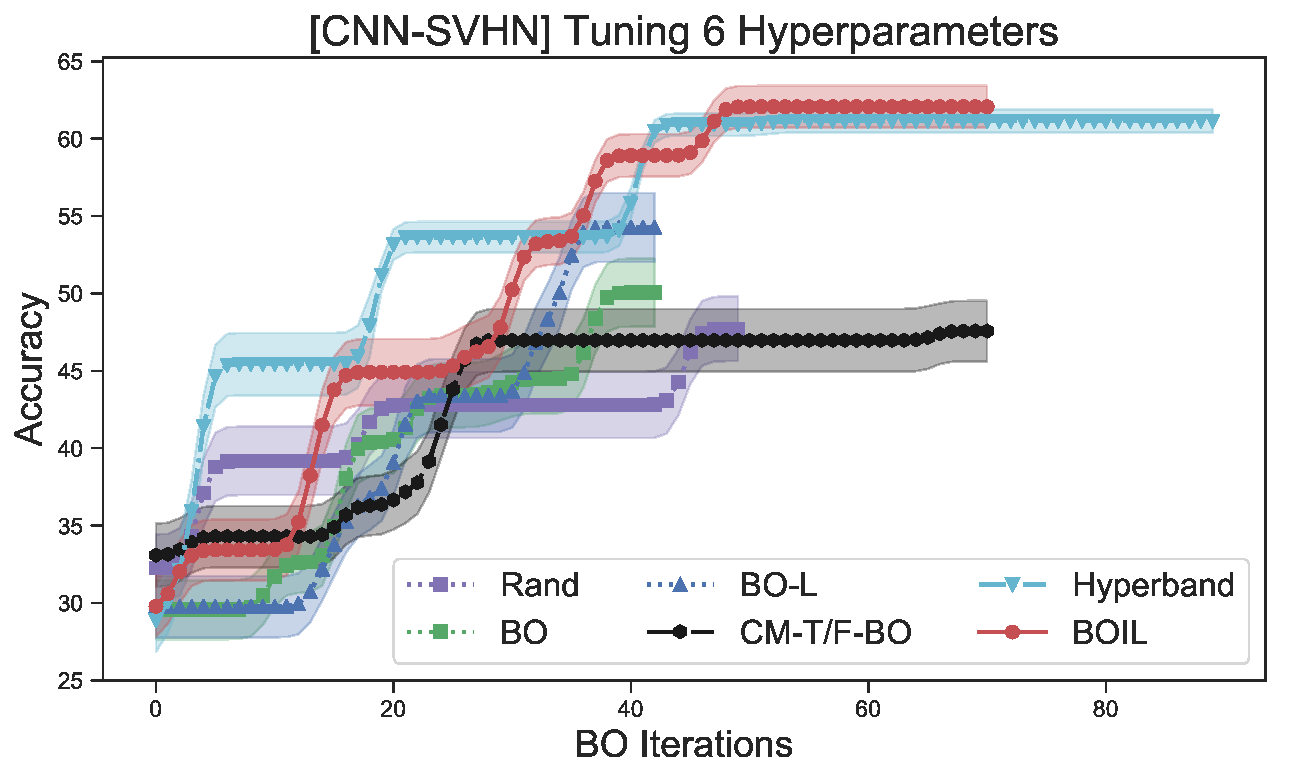}\includegraphics[width=0.5\columnwidth]{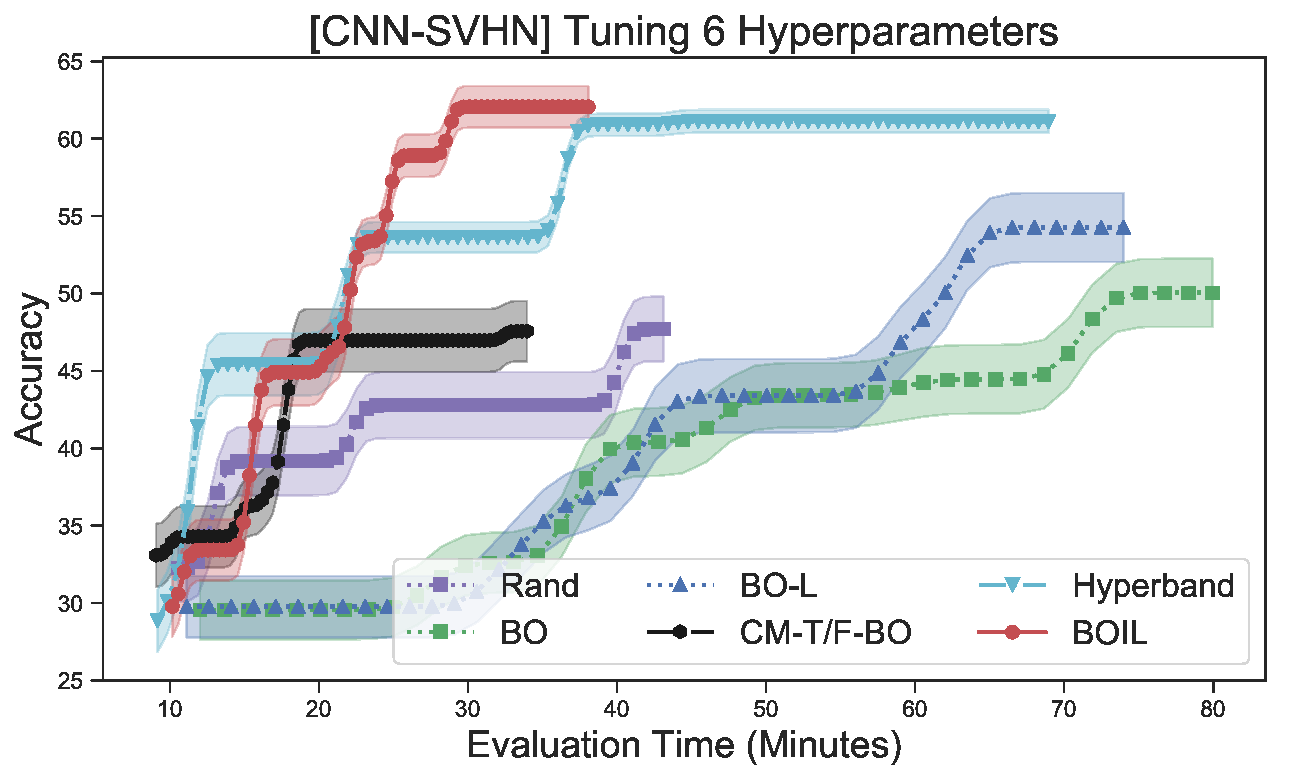}
\par\end{centering}
\caption{Comparison over BO evaluations (Left) and real-time (Right). Given
the same time budget, CM-T/F-BO, Hyperband and BOIL can take more
evaluations than vanilla BO, BO-L and Rand. BOIL outperforms other
competitors in finding the optimal parameters in an iteration-efficient
manner.   CM-T/F-BO does not augment the observations from the curve
and requires more evaluations. The results of InvertedPendulum
and CNN-CIFAR10 are in the appendix. \label{fig:Performance-comparison}}

\centering{}\vspace{-10pt}
\end{figure*}


\paragraph{Comparisons by iterations and real-time.}

Fig. \ref{fig:Performance-comparison} illustrates the performance
of different algorithms against the number of iterations as well as
real-time (the plots for CIFAR10 are in the appendix). The performance
is the utility score of the best hyperparameters identified by the
baselines. Across all three tasks, BOIL identifies optimal hyperparameters
using significantly less computation time than other approaches. 

The plots show that other approaches such as BO and BO-L can identify
well-performing hyperparameters in fewer iterations than BOIL. However,
they do so only considering costly, high-fidelity evaluations resulting
in significantly higher evaluation times. In contrast to this behavior,
BOIL accounts for the evaluation costs and chooses to initially evaluate
low-fidelity settings consuming less time. This allows fast assessments
of a multitude of hyperparameters. The information gathered here is
then used to inform later point acquisitions. Hereby, the inclusion
of augmented observations is crucial in offering useful information
readily available from the data. In addition, this augmentation is
essential to prevent from the GP kernel issue instead of adding the
full curve of points into our GP model.

Hyperband \cite{li2018hyperband} exhibits similar behavior in that
it uses low fidelity (small $t$) evaluations to reduce a pool of
randomly sampled configurations before evaluating at high fidelity
(large $t$). To deal with noisy evaluations and other effects, this
process is repeated several times. This puts Hyperband at a disadvantage
particularly in the noisy DRL tasks. Since early performance fluctuates
hugely, Hyperband can be misled in where to allocate evaluation effort.
It is then incapable of revising these choices until an entirely new pool of hyperparameters is sampled and evaluated from scratch. In contrast
to this, BOIL is more flexible than Hyperband in that it can freely
explore-exploit the whole joint space. The GP surrogate hereby allows
BOIL to generalize across hyperparameters and propagate information
through the joint space.

%% file: conclusion_v2.tex
Our framework complements the existing BO toolbox for hyperparameter
tuning with iterative learning. We present a way of leveraging our
understanding that later stages of the training process are informed
by progress made in earlier ones. This results in a more iteration-efficient
hyperparameter tuning algorithm that is applicable to a broad range
of machine learning systems. We evaluate its performance on a set
of diverse benchmarks. The results demonstrate that our model surpasses
the performance of well-established alternatives while consuming significantly
fewer resources. Finally, we note that our approach is not necessarily
specific to machine learning algorithms, but more generally applies
to any process exhibiting an iterative structure to be exploited.

%% file: NeurIPS2020_supplement.tex


\newpage

The following sections are intended to give the reader further insights
into our design choices and a deeper understanding of the algorithms
properties. First, we give a brief overview of Bayesian optimization
with Gaussian processes. We then illustrate our models behavior on
a two dimensional problem. Last, we give further details of our experiments
for reproducibility purposes.

\section{Bayesian Optimization Preliminaries}

Bayesian optimization is a sequential approach to global optimization
of black-box functions without making use of derivatives. It uses
two components: a learned surrogate model of the objective function
and an acquisition function derived from the surrogate for selecting
new points to inform the surrogate with. In-depth discussions beyond
our brief overview can be found in recent surveys \cite{Brochu_2010Tutorial,frazier2018tutorial,Shahriari_2016Taking}.

\paragraph{Notation.}

We summarize all of the notations used in our model in Table \ref{tab:Notation-List}
for ease of reading.

\subsection{Gaussian processes}

We present the GP surrogate model for the black-box function
$f$ \cite{Rasmussen_2006gaussian}. A GP defines a probability distribution
over functions $f$ under the assumption that any subset of points
$\left\{ (\bx_{i},f(\bx_{i})\right\} $ is normally distributed. Formally,
this is denoted as:
\begin{align*}
f(\bx)\sim \text{GP}\left(m\left(\bx\right),k\left(\bx,\bx'\right)\right),
\end{align*}
where $m\left(\bx\right)$ and $k\left(\bx,\bx'\right)$ are the mean
and covariance functions, given by $m(\bx)=\mathbb{E}\left[f\left(\bx\right)\right]$
and $k(\bx,\bx')=\mathbb{E}\left[(f\left(\bx\right)-m\left(\bx\right))(f\left(\bx'\right)-m\left(\bx'\right))^{T}\right]$.

Typically, the mean of the GP is assumed to be zero everywhere. The kernel
$k(\bx,\bx')$ can be thought of as a similarity measure relating
$f(\bx)$ and $f(\bx')$. Numerous kernels encoding different prior
beliefs about $f(\bx)$ have been proposed. A popular choice is given
by the square exponential kernel $k(\bx,\bx')=\sigma_{f}^{2}\exp\left[-(\bx-\bx')^{2}/2\sigma_{l}^{2}\right]$.
The length- and output-scales $\sigma_{l}^{2}$ and $\sigma_{f}^{2}$ regulate the maximal covariance
between two points and can be estimated using maximum marginal likelihood.
The SE kernel encodes the belief that nearby points are highly correlated
as it is maximized at $k(\bx,\bx)=\sigma_{f}^{2}$ and decays the
further $\bx$ and $\bx'$ are separated.

For predicting $f_{*}=f\left(\bx_{*}\right)$ at a new data point $\bx_{*}$,  assuming a zero mean $m(\bx)=0$, we have:
\begin{align}
\left[\begin{array}{c}
\boldsymbol{f}\\
f_{*}
\end{array}\right] & \sim\mathcal{N}\left(0,\left[\begin{array}{cc}
\bK & \bk_{*}^{T}\\
\bk_{*} & k_{**}
\end{array}\right]\right)\label{eq:p(f|f*)}
\end{align}
 where $k_{**}=k\left(\bx_{*},\bx_{*}\right)$, $\bk_{*}=[k\left(\bx_{*},\bx_{i}\right)]_{\forall i\le N}$
and $\bK=\left[k\left(\bx_{i},\bx_{j}\right)\right]_{\forall i,j\le N}$.
The conditional probability of $p\left(f_{*}\mid\boldsymbol{f}\right)$
follows a univariate Gaussian distribution as $p\left(f_{*}\mid\boldsymbol{f}\right)\sim\mathcal{N}\left(\mu\left(\bx_{*}\right),\sigma^{2}\left(\bx_{*}\right)\right)$.
Its mean and variance are given by:
\begin{align*}
\mu\left(\bx_{*}\right)= & \mathbf{k}_{*}\mathbf{K}^{-1}\mathbf{y}\\
\sigma^{2}\left(\bx_{*}\right)= & k_{**}-\mathbf{k}_{*}\mathbf{K}^{-1}\mathbf{k}_{*}^{T}.
\end{align*}
 As GPs give full uncertainty information with any prediction, they
provide a flexible nonparametric prior for Bayesian optimization.
We refer the interested readers to \cite{Rasmussen_2006gaussian}
for further details on GPs.
\begin{table*}
\caption{Notation List\label{tab:Notation-List}}

\centering{}%
\begin{tabular}{ccc}
\toprule 
Parameter & Domain & Meaning\tabularnewline
\midrule
$d$ & integer, $\mathcal{N}$ & dimension, no. of hyperparameters to be optimized\tabularnewline
$\bx$ & vector,$\mathcal{R}^{d}$ & input hyperparameter\tabularnewline
$N$ & integer, $\mathcal{N}$ & maximum number of BO iterations\tabularnewline
$T_{\textrm{min}}$, $T_{\textrm{max}}$ & integer, $\mathcal{N}$ & the min/max no of iterations for training a ML algorithm\tabularnewline
$t$ & $\in [ T_{\textrm{min}},...T_{\textrm{max}}]$ & index of training steps\tabularnewline
$M$ & integer, $\mathcal{N}$ & the maximum number of augmentation. We set $M=15$.\tabularnewline
$\delta$ & scalar, $\mathcal{R}$ & threshold for rejecting augmentation when ln of cond$(K)>\delta$
\tabularnewline
$m$ &  $\in \{1,...M\}$ & index of augmenting variables\tabularnewline
$n$ &  $\in \{1,...,N\}$ & index of BO iterations\tabularnewline
$\bz=[\bx,t]$ & vector, $\mathcal{R}^{d+1}$ & concatenation of the parameter $\bx$ and iteration
$t$\tabularnewline
$c_{n,m}$ & scalar, $\mathcal{R}$ & training cost (sec)\tabularnewline
$y_{n}$ & scalar, $\mathcal{R}$ & transformed score at the BO iteration $n$\tabularnewline
$y_{n,m}$ & scalar, $\mathcal{R}$ & transformed score at the BO iteration $n$, training step $m$\tabularnewline
$\alpha(\bx,t)$ & function & acquisition function for performance\tabularnewline
$\mu_{c}(\bx,t)$ & function & estimation of the cost by LR given $\bx$ and $t$\tabularnewline
$r(.\mid \bx,t)$ & function & a raw learning curve, $r(\bx,t)=[r(1\mid\bx,t),...r(t'\mid\bx,t),r(t\mid\bx,t)]$\tabularnewline
$f(\bx,t)$ & function & a black-box function which is compressed from the above $f()$\tabularnewline
$l\left(.\mid m_{0},g_{0}\right)$ & function & Logistic curve $l(u \mid m_{0},g_{0})=\frac{1}{1+\exp\left(-g_{0}\left[u-m_{0}\right]\right)}$\tabularnewline
$g_{0}$, $g_{0}^{*}$ & scalar, $\mathcal{R}$ & a growth parameter defining a slope, $g_{0}^{*}=\arg\max_{g_{0}}L$\tabularnewline
$m_{0}$, $m_{0}^{*}$ & scalar, $\mathcal{R}$ & a middle point parameter, $m_{0}^{*}=\arg\max_{m_{0}}L$\tabularnewline
$L$ & scalar, $\mathcal{R}$ & Gaussian process log marginal likelihood\tabularnewline
\end{tabular}
\end{table*}

\subsection{Acquisition function}

Bayesian optimization is typically applied in settings in which the
objective function is expensive to evaluate. To minimize interactions
with that objective, an acquisition function is defined to reason
about the selection of the next evaluation point $\bx_{t+1}=\arg\max_{x\in\mathcal{X}}\alpha_{t}\left(\bx\right)$.
The acquisition function is constructed from the predictive mean and
variance of the surrogate to be easy to evaluate and represents the
trade-off between exploration (of points with high predictive uncertainty)
and exploitation (of points with high predictive mean). Thus, by design
the acquisition function can be maximized with standard global optimization
toolboxes. Among the many acquisition functions \cite{Hennig_2012Entropy,Hernandez_2014Predictive,Jones_1998Efficient,ru2018fast,Srinivas_2010Gaussian,Wang_2017Max}
 available in the literature, the expected improvement \cite{Jones_1998Efficient,Nguyen_ACML2017Regret,Wang_2014Theoretical}
is one of the most popular.

\subsection{GP kernels and treatment of GP hyperparameters}

We present the GP kernels and treatment of GP hyperparameters for
the black-box function $f$.

Although the raw learning curve in DRL is noisy, the transformed version
using our proposed curve compression is smooth.
Therefore, we use two squared exponential kernels for input hyperparameter
and training iteration, respectively. That is $k_{x}(\bx,\bx')=\exp\left(-\frac{||\bx-\bx'||^{2}}{2\sigma_{x}^{2}}\right)$
and $k_{t}(t,t')=\exp\left(-\frac{||t-t'||^{2}}{2\sigma_{t}^{2}}\right)$
where the observation $\bx$ and $t$ are normalized to $[0,1]^{d}$
and the outcome $y$ is standardized $y\sim\mathcal{N}\left(0,1\right)$
for robustness. As a result, our product kernel becomes
\begin{align*}
k\left([\bx,t],[\bx',t']\right) & =k(\bx,\bx')\times k(t,t')=\exp\left(-\frac{||\bx-\bx'||^{2}}{2\sigma_{x}^{2}}-\frac{||t-t'||^{2}}{2\sigma_{t}^{2}}\right).
\end{align*}

The length-scales $\sigma_{x}$ and $\sigma_{t}$ are learnable parameters indicating the variability of the function with regards to the hyperparameter input $\bx$ and number of training iterations $t$. Estimating appropriate values for them is critical as this represents the GPs prior regarding the sensitivity of performance w.r.t. changes in the number of training iterations and hyperparameters. For extremely large $\sigma_{t}$ we expect the objective function to change very little for different numbers of training iterations. For small $\sigma_{t}$ by contrast we expect drastic changes even for small differences. We estimate these GP hyperparameters (including the length-scales $\sigma_{x}$, $\sigma_{t}$ and the output noise variance $\sigma_y$) by maximizing their log marginal likelihood \cite{Rasmussen_2006gaussian}.


We optimize Eq. (\ref{eq:MarginalLLK}) with a gradient-based optimizer, providing the analytical gradient to the algorithm. We start the optimization from the previous hyperparameter values $\theta_{prev}$.  If the optimization fails due to numerical issues, we keep the previous value of the hyperparameters. We refit the hyperparameters every $3\times d$ function evaluations where $d$ is the dimension.

\begin{figure*}
\begin{centering}
\includegraphics[width=1\columnwidth]{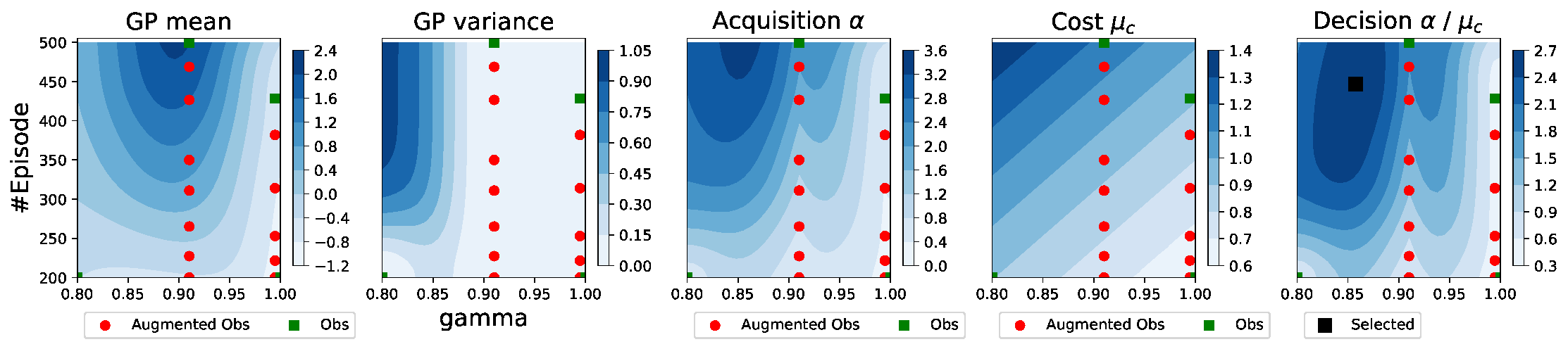}
\par\end{centering}
\begin{centering}
\includegraphics[width=1\columnwidth]{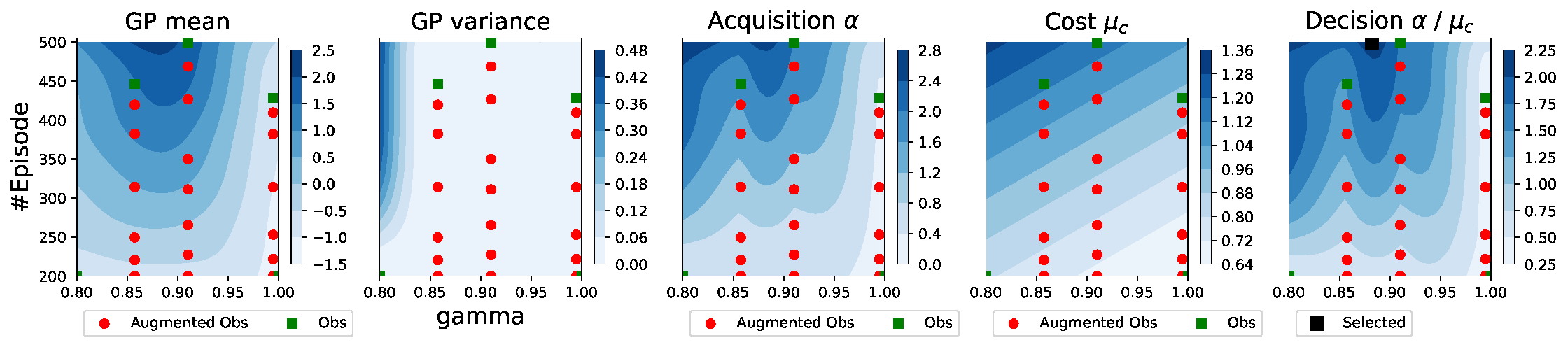}
\par\end{centering}
\begin{centering}
\includegraphics[width=1\columnwidth]{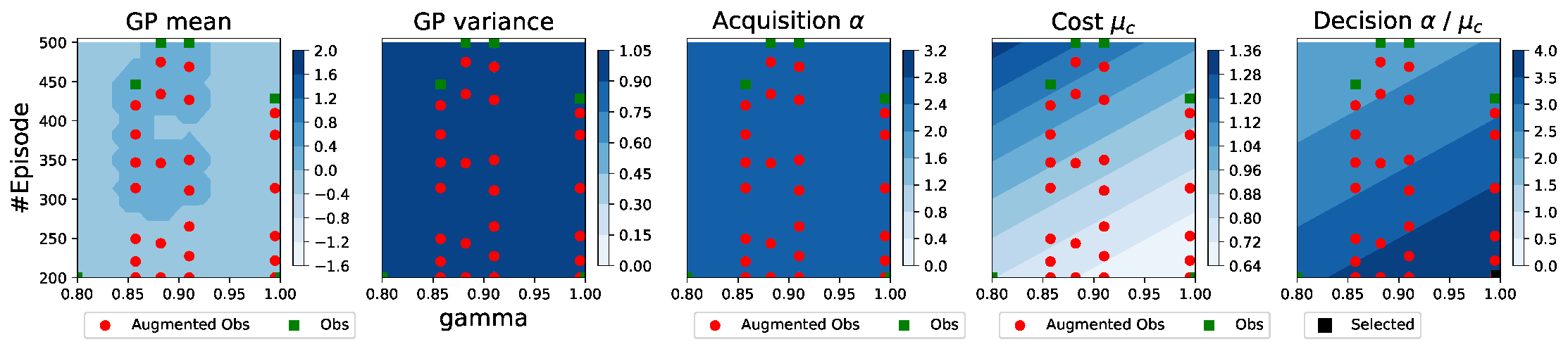}
\par\end{centering}
\begin{centering}
\includegraphics[width=1\columnwidth]{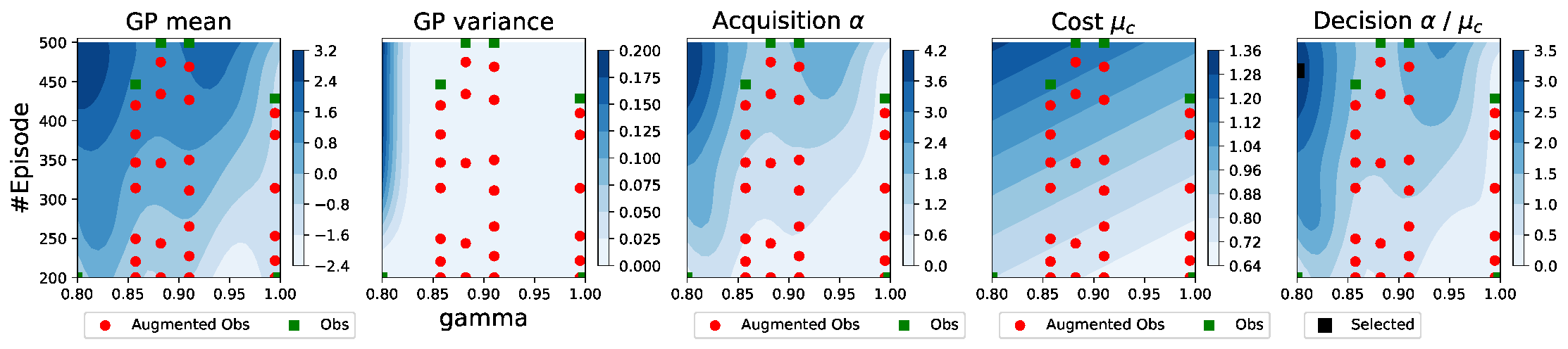}
\par\end{centering}
\begin{centering}
\includegraphics[width=1\columnwidth]{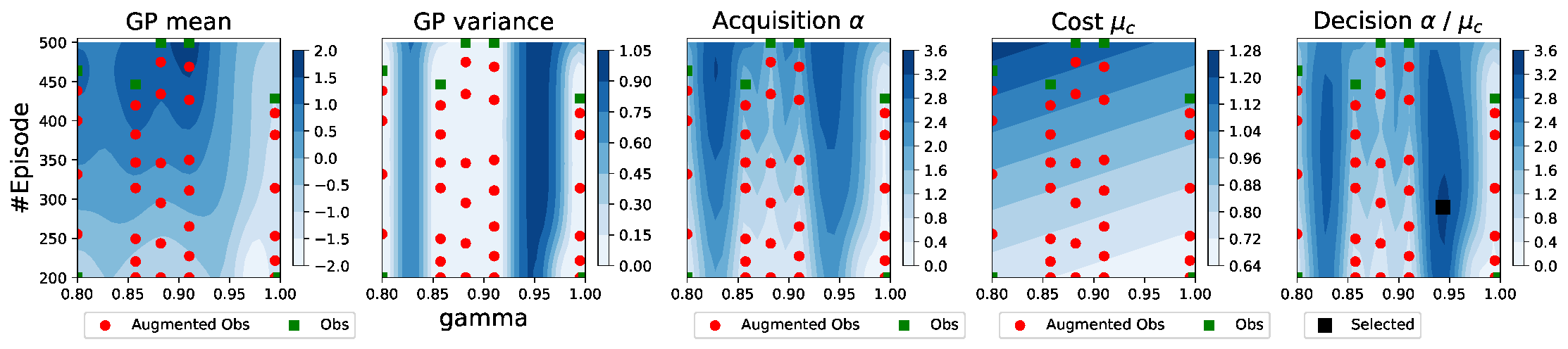}
\par\end{centering}
\caption{Illustration of BOIL on a $2$-dimensional optimization task of DDQN
on CartPole. The augmented observations fill the joint hyperparameter-iteration
space quickly to inform our surrogate. Our decision balances utility
$\alpha$ against cost $\tau$ for iteration-efficiency. Especially
in situations of multiple locations sharing the same utility value,
our algorithm prefers to select the cheapest option. \label{fig:Illustration-our-BOIL}}
\end{figure*}

\begin{figure*}
\begin{centering}
\includegraphics[width=1\columnwidth]{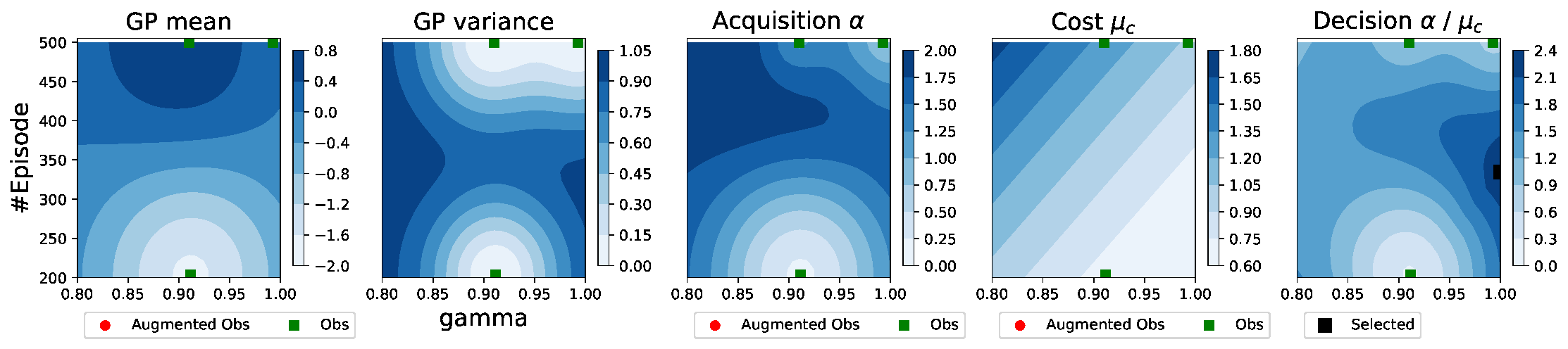}
\par\end{centering}
\begin{centering}
\includegraphics[width=1\columnwidth]{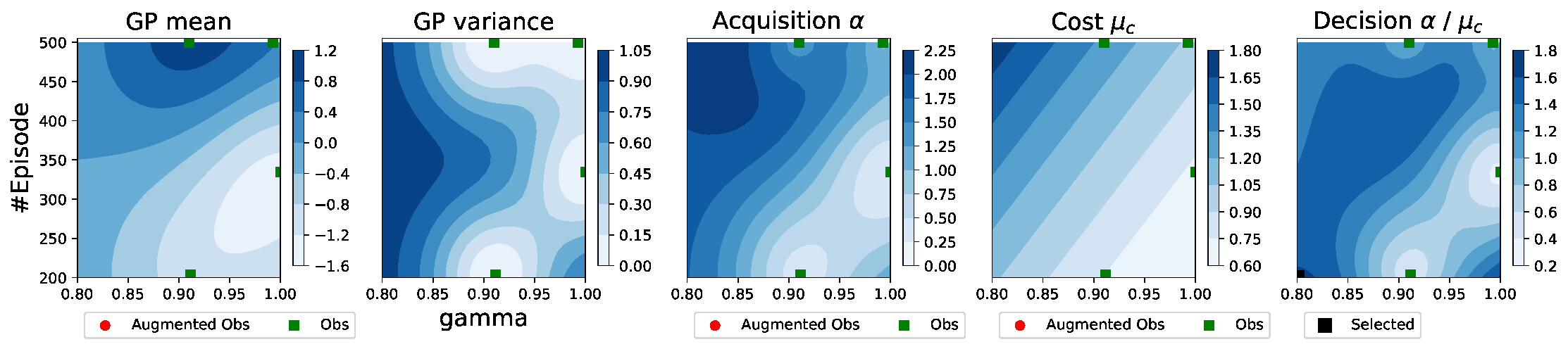}
\par\end{centering}
\begin{centering}
\includegraphics[width=1\columnwidth]{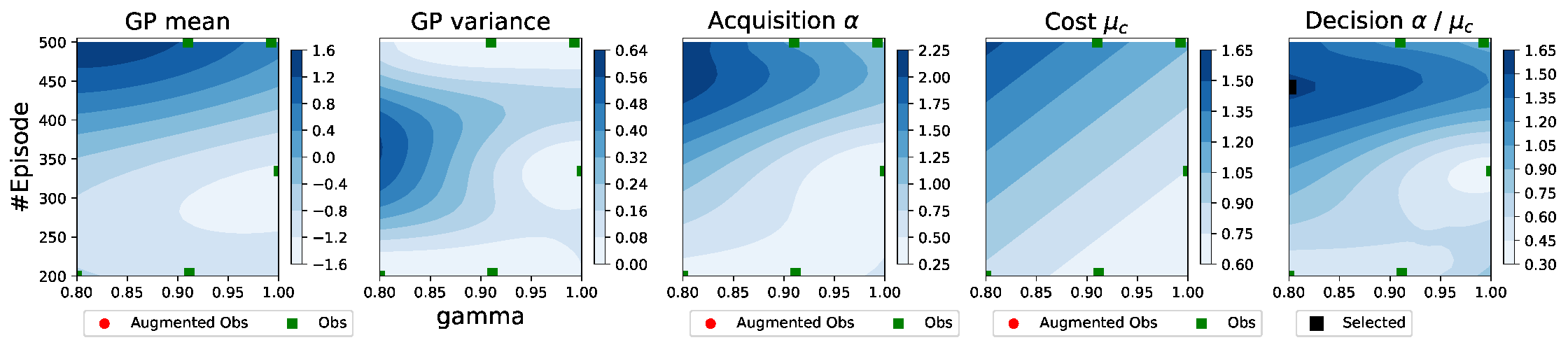}
\par\end{centering}
\begin{centering}
\includegraphics[width=1\columnwidth]{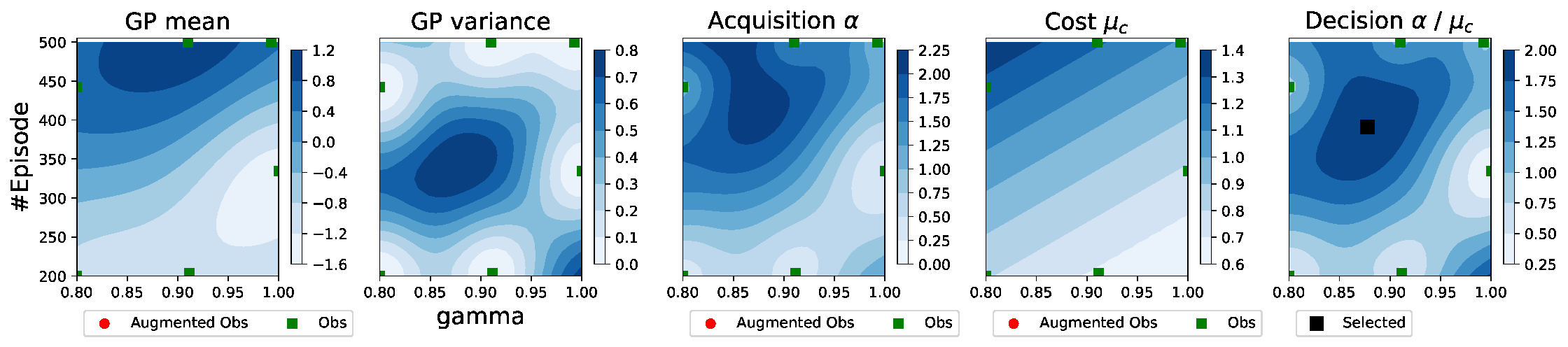}
\par\end{centering}
\begin{centering}
\includegraphics[width=1\columnwidth]{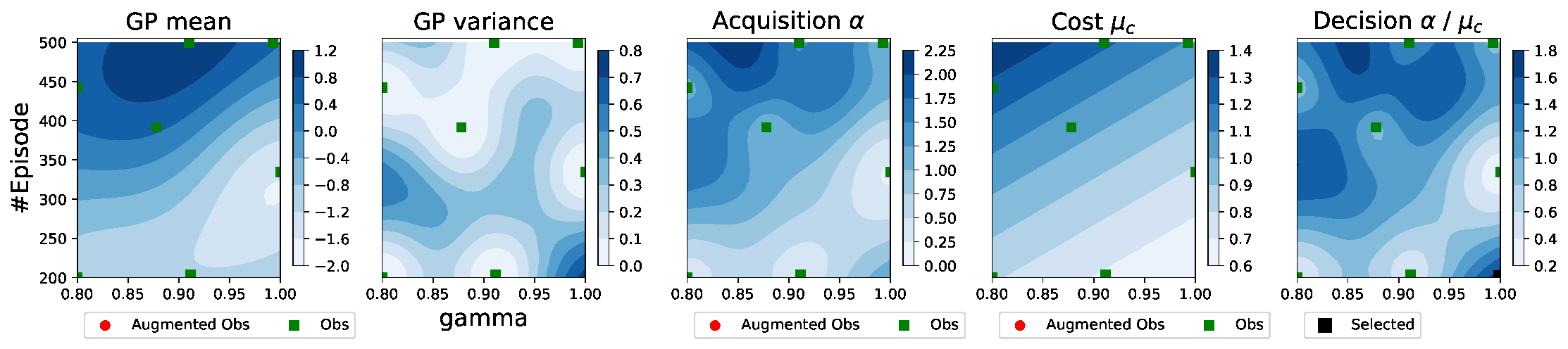}
\par\end{centering}
\begin{centering}
\caption{Illustration of the Continuous Multi task/fidelity BO (CM-T/F-BO)
-{}- this is the case of BOIL \textbf{without} using augmented observations
(same setting as Fig. \ref{fig:Illustration-our-BOIL}). This version
leads to less efficient optimization as the additional iteration dimension
requires more evaluation than optimizing the hyperparameters on their
own. \label{fig:Illustration-BOIML_noaugmented}
}
\par\end{centering}
\end{figure*}

\section{Algorithm Illustration and Further Experiments}

\begin{table}
\centering{}\caption{Dueling DQN algorithm on CartPole problem.\label{tab:Hyperparameters-of-DDQN}}
\begin{tabular}{cccc}
\toprule 
Variables & Min & Max & Best Found $\bx^{*}$\tabularnewline
\midrule
$\gamma$ discount factor & $0.8$ & $1$ & $0.95586$\tabularnewline
learning rate model & $1e^{-6}$ & $0.01$ & $0.00589$\tabularnewline
\#Episodes & $300$ & $800$ & -\tabularnewline
\bottomrule
\end{tabular}
\end{table}
Fig. \ref{fig:Illustration-our-BOIL} and Fig. \ref{fig:Illustration-BOIML_noaugmented}
illustrate the behavior of our proposed algorithm BOIL on the example
of optimizing the discount factor $\gamma$ of Dueling DQN \cite{wang2016dueling}
on the CartPole problem. The two settings differ in the inclusion
augmented observations into BOIL in Fig. \ref{fig:Illustration-our-BOIL}
and CM-T/F-BO (or BOIL without augmented observations) in Fig. \ref{fig:Illustration-BOIML_noaugmented}.

In both cases, we plot the GP predictive mean in Eq. (1), GP predictive
variance in Eq. (2), the acquisition function in Eq. (3), the predicted
function and the final decision function in Eq. (8). These equations
are defined in the main manuscript.
\begin{table}
\centering{}\caption{A2C algorithm on Reacher (left) and InvertedPendulum (right).\label{tab:Hyperparameters-of-Reacher}}
\begin{tabular}{cccc}
\toprule 
Variables & Min & Max & Best Found $\bx^{*}$\tabularnewline
\midrule
$\gamma$ discount factor & $0.8$ & $1$ & $0.8$\tabularnewline
learning rate actor & $1e^{-6}$ & $0.01$ & $0.00071$\tabularnewline
learning rate critic  & $1e^{-6}$ & $0.01$ & $0.00042$\tabularnewline
\#Episodes & $200$ & $500$ & -\tabularnewline
\bottomrule
\end{tabular}%
\begin{tabular}{cccc}
\toprule 
 & Min & Max & Best Found $\bx^{*}$\tabularnewline
\midrule
 & $0.8$ & $1$ & $0.95586$\tabularnewline
 & $1e^{-6}$ & $0.01$ & $0.00589$\tabularnewline
 & $1e^{-6}$ & $0.01$ & $0.00037$\tabularnewline
 & $700$ & $1500$ & -\tabularnewline
\bottomrule
\end{tabular}
\end{table}

As shown in the respective figures the final decision function balances
between utility and cost of any pair $(\gamma,t)$ to achieve iteration
efficiency. Especially in situations where multiple locations share
the same utility value, our decision will prefer to select the cheapest
option. Using the augmented observations in Fig. \ref{fig:Illustration-our-BOIL},
our joint space is filled quicker with points and the uncertainty
(GP variance) across it reduces faster than in Fig. \ref{fig:Illustration-BOIML_noaugmented}
-- the case of vanilla CM-T/F-BO without augmenting observations.
A second advantage of having augmented observations is that the algorithm
is discouraged to select the same hyperparameter setting at lower
fidelity than a previous evaluation. We do not add the full curve as it can be redundant while causing the conditioning problem of the GP covariance matrix.

\subsection{Experiment settings}

We summarize the hyperparameter search ranges for A2C on Reacher and InvertedPendulum in Table \ref{tab:Hyperparameters-of-Reacher}, CNN on SHVN in Table \ref{tab:Hyper-parameterCNN} and DDQN on CartPole in Table \ref{tab:Hyperparameters-of-DDQN}. Additionally, we present the best found parameter $\bx^{*}$ for these problems. Further details of the DRL agents are listed in Table \ref{tab:fixed_hyperparams}.

\begin{table}
\centering{}\caption{Convolutional Neural Network.\label{tab:Hyper-parameterCNN}}
\begin{tabular}{cccc}
\toprule 
Variables & Min & Max & Best Found $\bx^{*}$\tabularnewline
\midrule
filter size & $1$ & $8$ & $5$\tabularnewline
pool size & $1$ & $5$ & $5$\tabularnewline
batch size & $16$ & $1000$ & $8$\tabularnewline
learning rate & $1e^{-6}$ & $0.01$ & $0.000484$\tabularnewline
momentum & $0.8$ & $0.999$ & $0.82852$\tabularnewline
decay & $0.9$ & $0.999$ & $0.9746$\tabularnewline
number of epoch & $30$ & $150$ & -\tabularnewline
\bottomrule
\end{tabular}
\end{table}

\begin{table}

\caption{Further specification for DRL agents\label{tab:fixed_hyperparams}}

\begin{centering}
\begin{tabular}{cc}
\toprule 
Hyperparameter & Value\tabularnewline
\midrule
\multicolumn{2}{c}{A2C}\tabularnewline
Critic-network architecture & $[32,32]$\tabularnewline
Actor-network architecture & $[32,32]$\tabularnewline
Entropy coefficient & $0.01$\tabularnewline
\end{tabular}%
\begin{tabular}{cc}
\toprule 
\multicolumn{2}{c}{Dueling DQN}\tabularnewline
Q-network architecture & $[50,50]$\tabularnewline
\textgreek{e}-greedy (start, final, number of steps) & $(1.0,0.05,10000)$\tabularnewline
Buffer size & $10000$\tabularnewline
Batch size & $64$\tabularnewline
PER-$\alpha$ \cite{schaul2015prioritized} & $1.0$\tabularnewline
PER-$\beta$ (start, final, number of steps) & $(1.0,0.6,1000)$\tabularnewline
\bottomrule
\end{tabular}
\par\end{centering}

\end{table}

\begin{figure*}
\includegraphics[width=0.5\columnwidth]{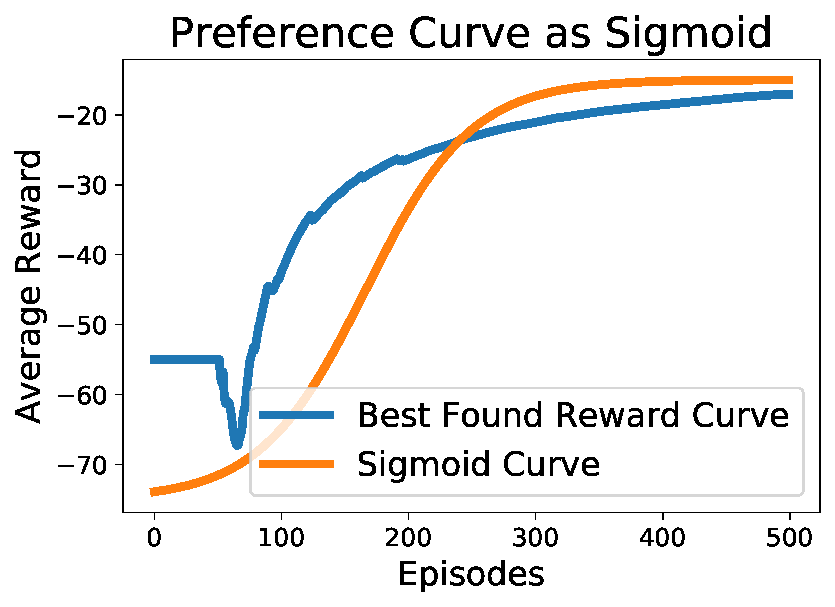}\includegraphics[width=0.5\columnwidth]{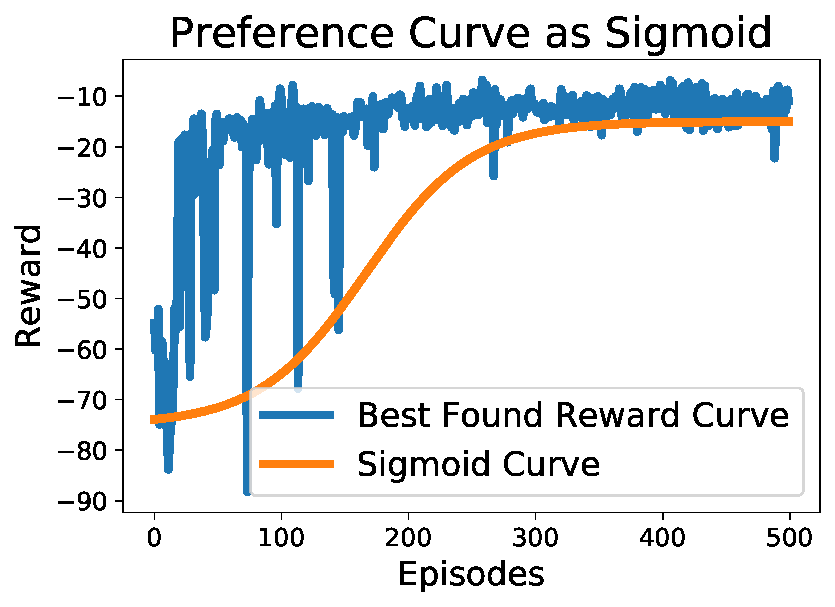}

\includegraphics[width=0.5\columnwidth]{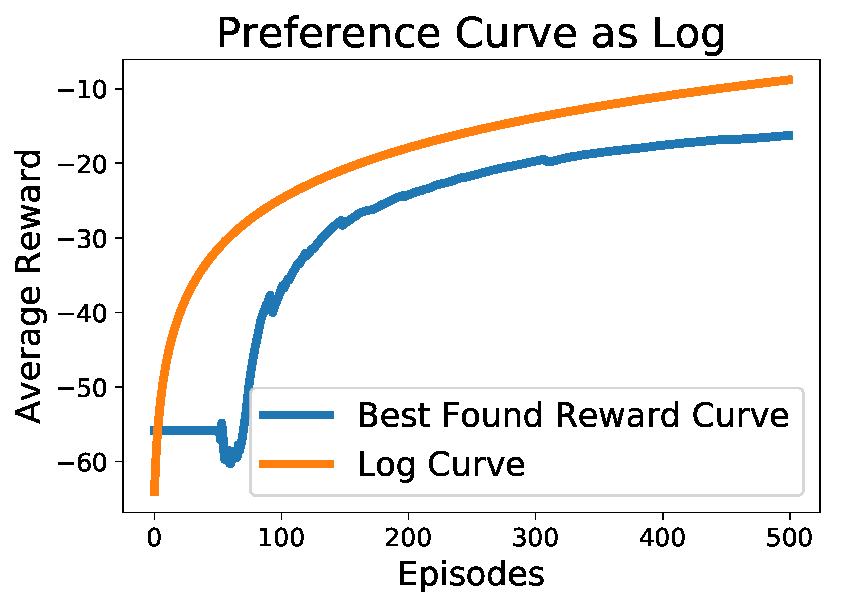}\includegraphics[width=0.5\columnwidth]{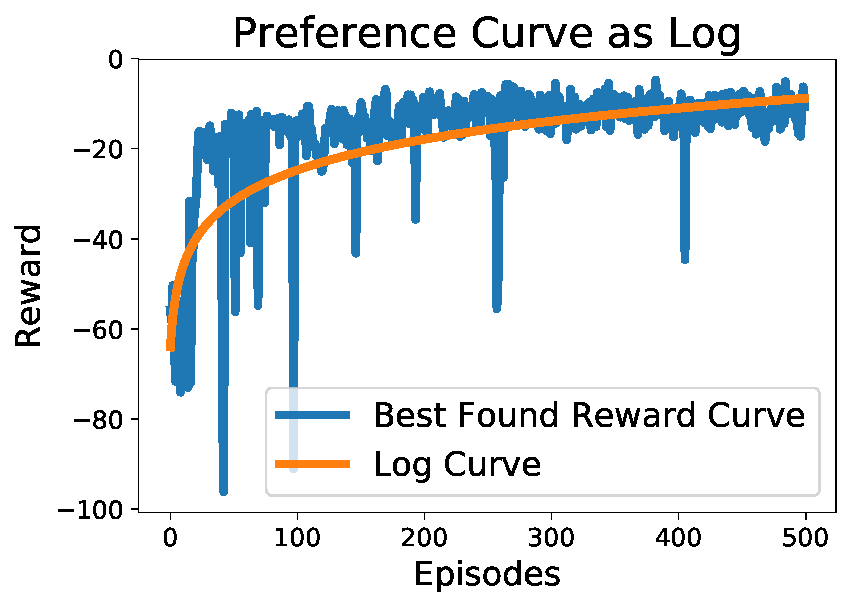}

\includegraphics[width=0.5\columnwidth]{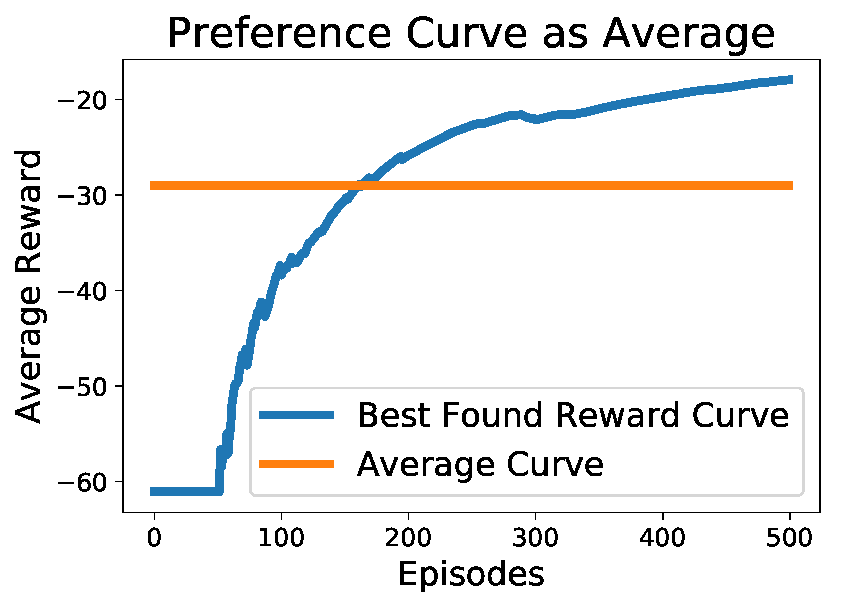}\includegraphics[width=0.5\columnwidth]{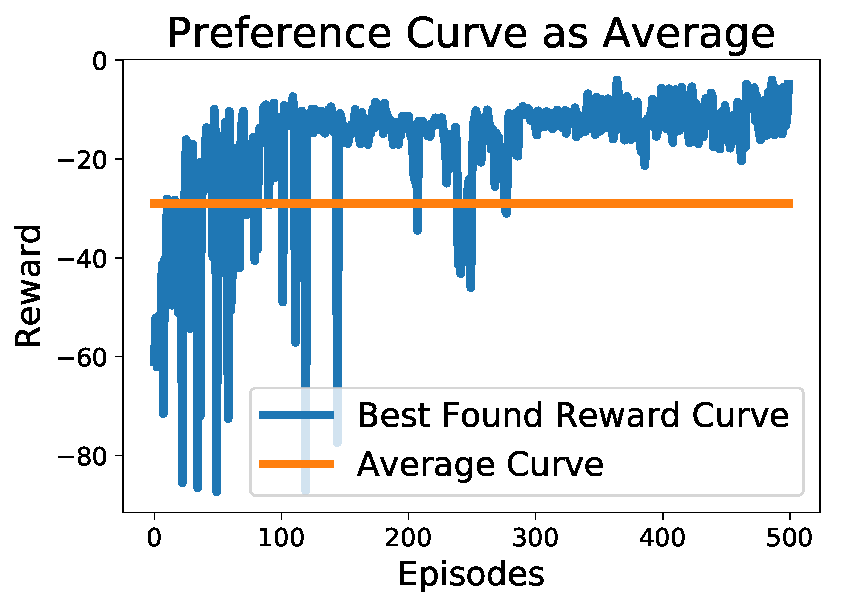}

\caption{To highlight the robustness, we examine the results using different
preference functions such as Sigmoid curve, Log curve, and Average
curve on Reacher experiments. The results include the best found reward
curve with different preference choices that show the robustness of
our model. Left column: the best found curve using averaged reward
over 100 consecutive episodes. Right column: the best found curve
using the original reward. \label{fig:experiment_different_pref}}
\end{figure*}

\subsection{Learning Logistic Function}

\begin{figure}
\begin{centering}
\includegraphics[width=0.92\textwidth]{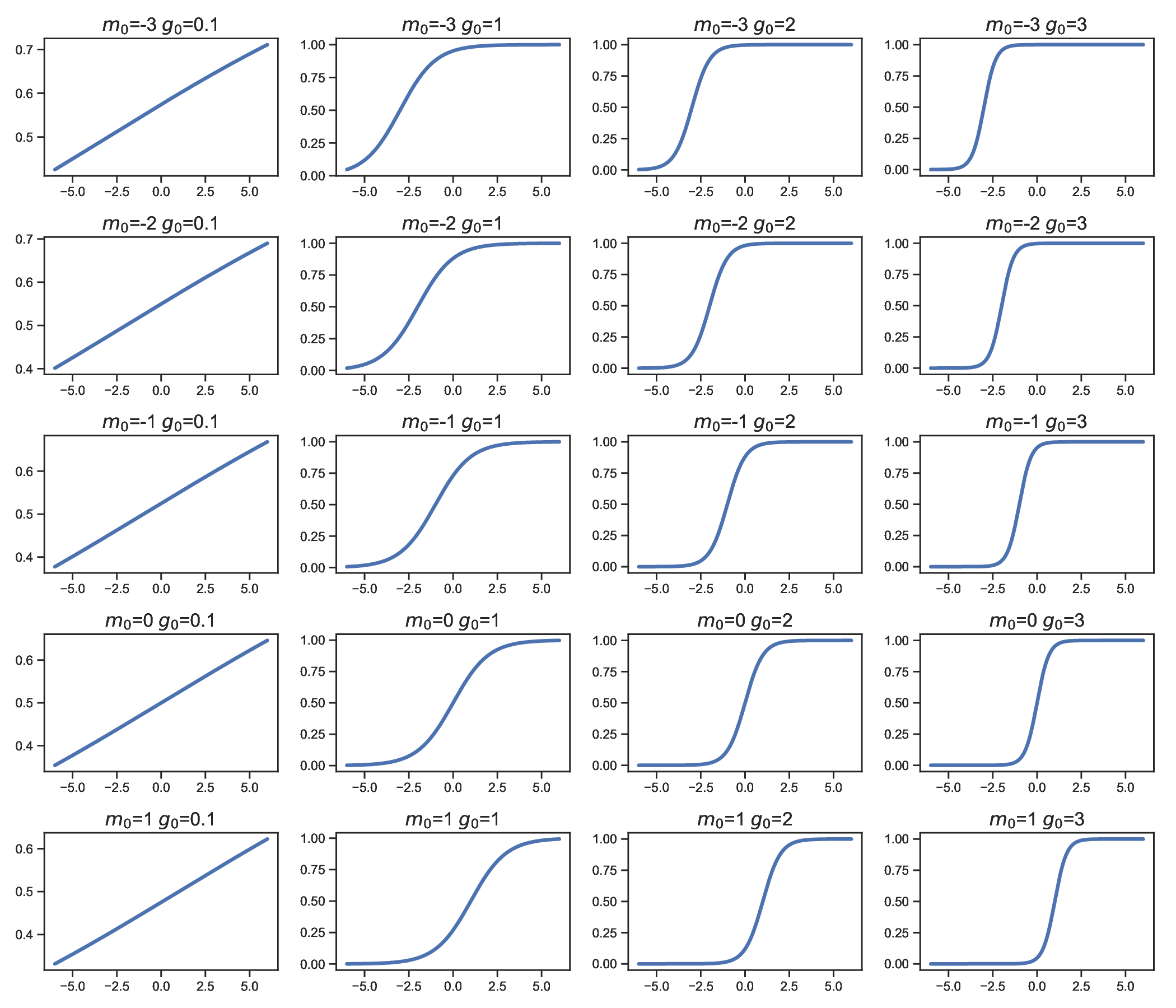}
\par\end{centering}
\caption{Examples of Logistic function $l(u)=\frac{1}{1+\exp\left(-g_{0}\left[u-m_{0}\right]\right)}$
with different values of middle parameter $m_{0}$ and growth parameter
$g_{0}$.\label{fig:Examples-of-Logistic}}

\end{figure}

\begin{figure}
\centering
\includegraphics[width=0.4\textwidth]{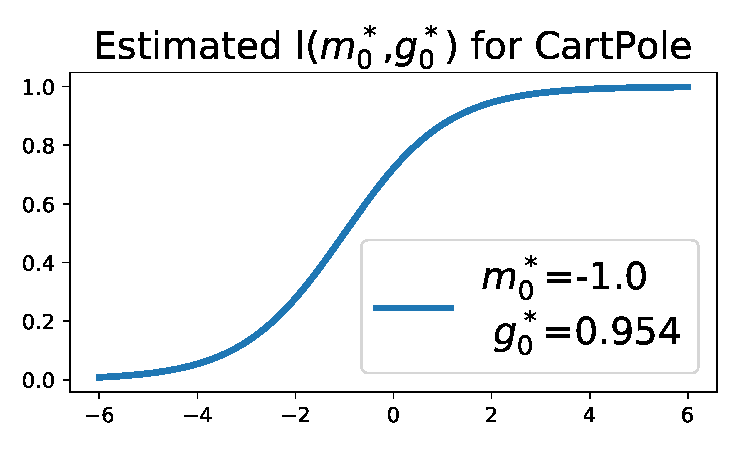}\includegraphics[width=0.4\textwidth]{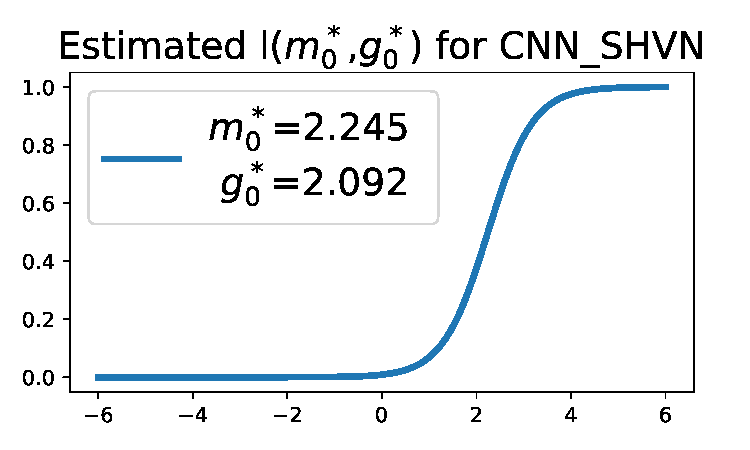}
\centering
\includegraphics[width=0.4\textwidth]{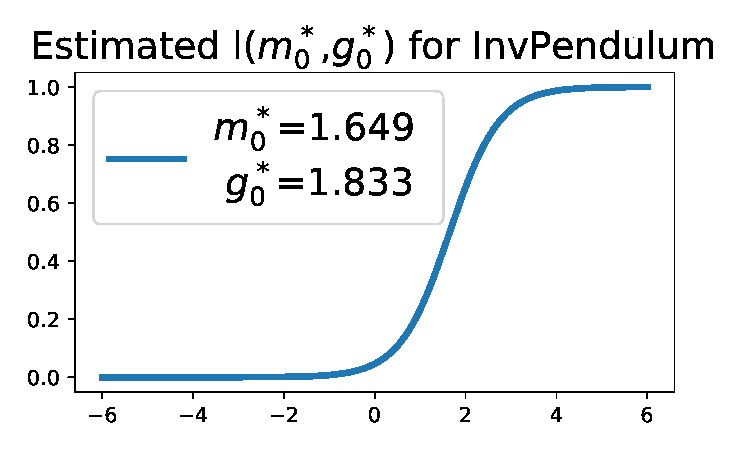}\includegraphics[width=0.4\textwidth]{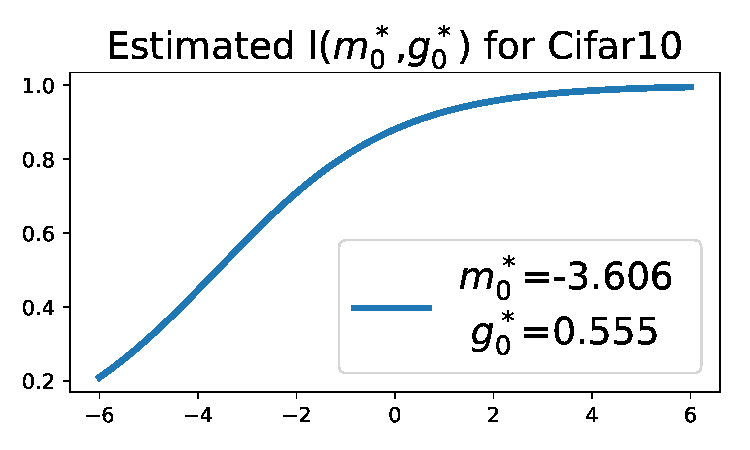}
\centering
\includegraphics[width=0.4\textwidth]{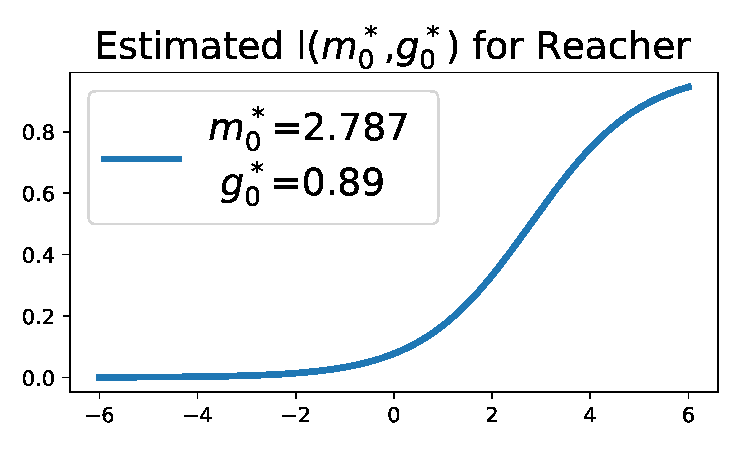}

\caption{We learn the suitable transformation curve directly from the data.
We parameterized the Logistic curve as $l\left(m_{0},g_{0}\right)=\frac{1}{1+\exp\left(-g_{0}\left[1-m_{0}\right]\right)}$
then estimate $g_{0}$ and $m_{0}$. The estimated function $l(m_{0}^{*},g_{0}^{*})$
is then used to compress our curve. The above plots are the estimated
$l()$ at different environments and datasets.\label{fig:EstimatingLogisticCurve}}

\end{figure}

We first present the Logistic curve $l(u \mid \bx, t)=\frac{1}{1+\exp\left(-g_{0}\left[u-m_{0}\right]\right)}$
using different choices of $g_{0}$ and $m_{0}$ in Fig. \ref{fig:Examples-of-Logistic}.
We then learn from the data to get the optimal choices $g_{0}^{*}$
and $m_{0}^{*}$ presented in Fig. \ref{fig:EstimatingLogisticCurve}.

\subsection{Robustness over Different Preference Functions}

\begin{figure*}
\includegraphics[width=0.49\columnwidth]{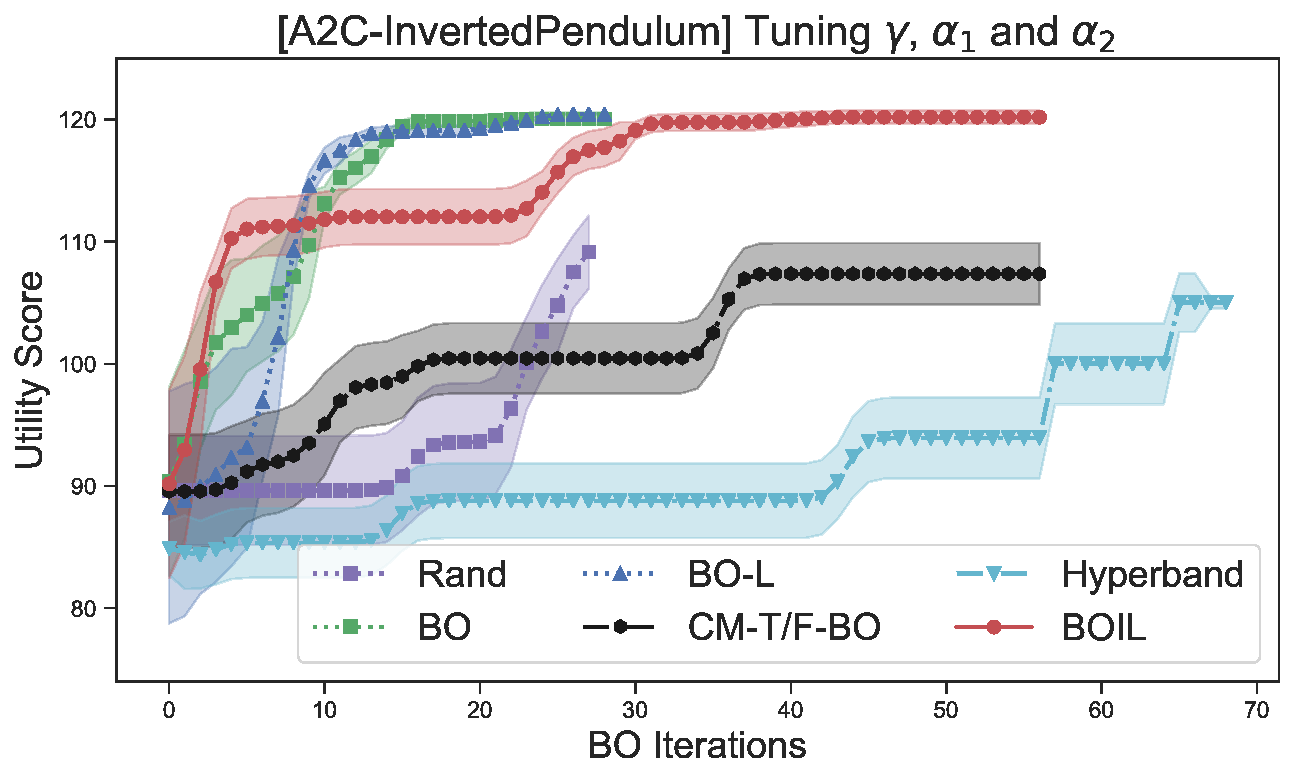}\includegraphics[width=0.49\columnwidth]{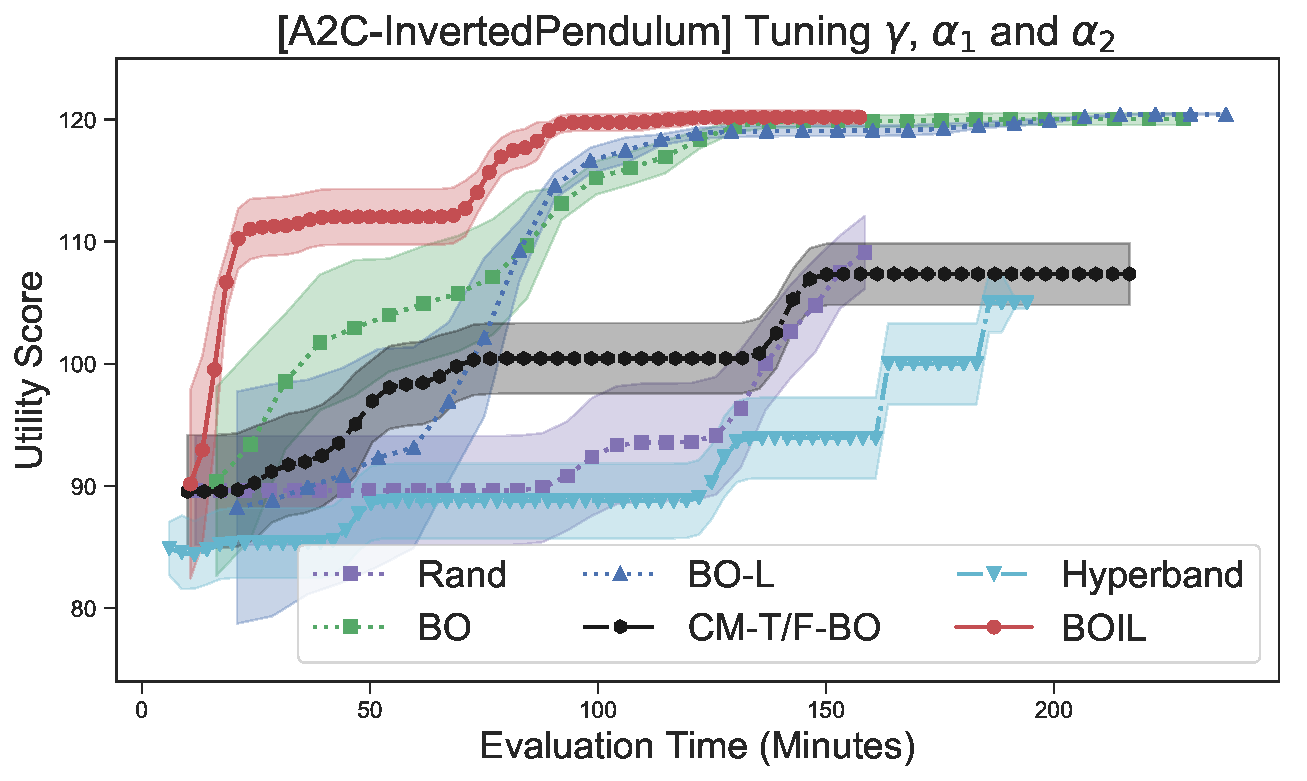}

\includegraphics[width=0.5\columnwidth]{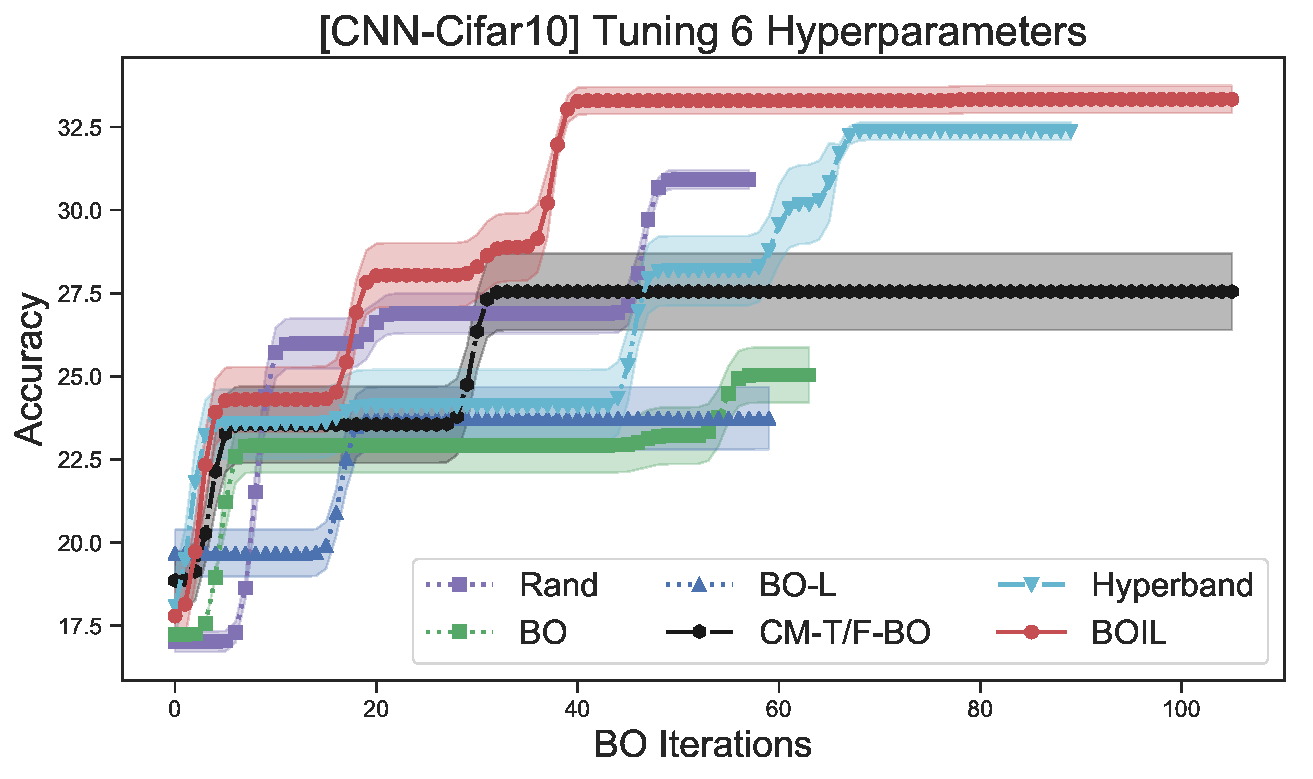}\includegraphics[width=0.5\columnwidth]{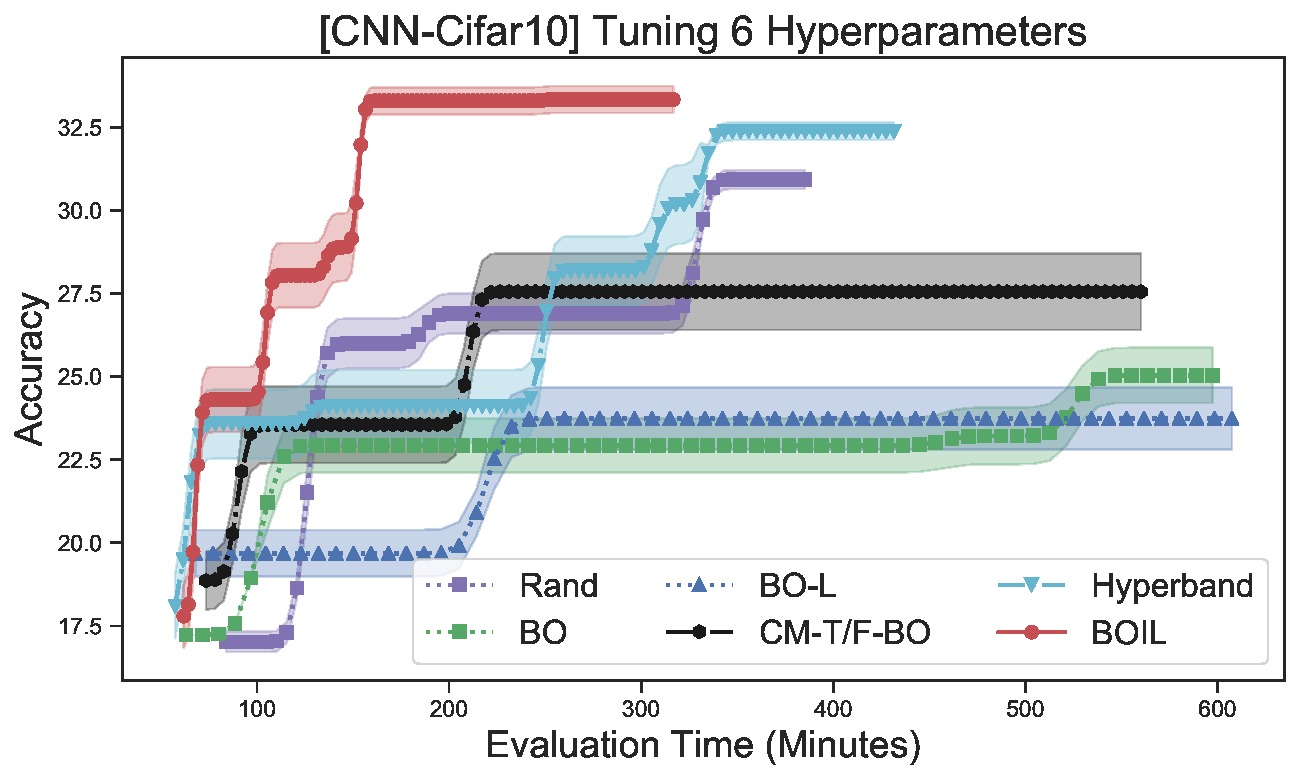}

\caption{Tuning hyperparameters of a DRL on InvertedPendulum and a CNN model
on CIFAR10.\label{fig:CIFAR10}}
\end{figure*}
We next study the learning effects with respect to different choices
of the preference functions. We pick three preference functions including
the Sigmoid, Log and Average to compute the utility score for each
learning curve. Then, we report the best found reward curve under
such choices. The experiments are tested using A2C on Reacher-v2.
The results presented in Fig. \ref{fig:experiment_different_pref}
demonstrate the robustness of our model with the preference functions.

\subsection{Applying Freeze-Thaw BO in the settings considered}
\begin{figure}
    \centering
    \includegraphics[width=0.99\linewidth]{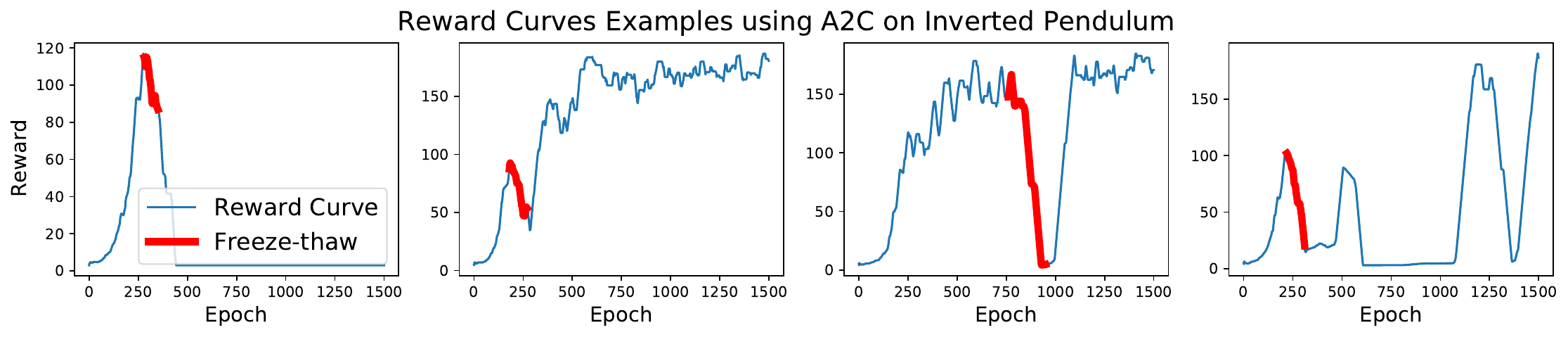}
    \caption{Illustration of Freeze-thaw BO in DRL. Freeze-thaw BO will terminate training processes when  training performance (in \textcolor{blue}{blue}) significantly drops (i.e. at the \textcolor{red}{red} locations) as the \textit{exponential decay} model will predict low final performance. In most RL enviroments noisy training curves are unavoidable. Thus, Freeze-thaw BO will dismiss all curves including good setting, never completing a single training run before the final epoch.}  \label{fig:freezethaw}
\end{figure}

While both the exponential decay in Freeze-Thaw BO \cite{swersky2014freeze} and our compression function encode preferences regarding training development, there is an important distinction between the two approaches. Freeze-thaw BO utilises the exponential decay property to \textit{terminate} the training curve, while BOIL only uses the sigmoid curve to \textit{guide} the search. We refer to Fig. \ref{fig:freezethaw} for further illustration  of why Freeze-thaw BO struggles in DRL settings.

\subsection{Ablation Study using Freeze-Thraw Kernel for Time}

In the joint modeling framework of hyperparameter and time (iteration),
we can replace the kernel either $k(\bx,\bx)$ or $k(t,t)$ with different
choices. We, therefore, set up a new baseline of using the time-kernel
$k(t,t')$ in Freeze-Thaw approach \cite{swersky2014freeze} which
encodes the monotonously exponential decay from the curve. Particularly,
we use the kernel defined as
\begin{align*}
k(t,t') & =\frac{\beta^{\alpha}}{\left(t+t'+\beta\right)^{\alpha}}
\end{align*}
for parameters $\alpha,\beta>0$ which are optimized in the GP models.

\begin{figure}
\begin{centering}
\includegraphics[width=0.5\textwidth]{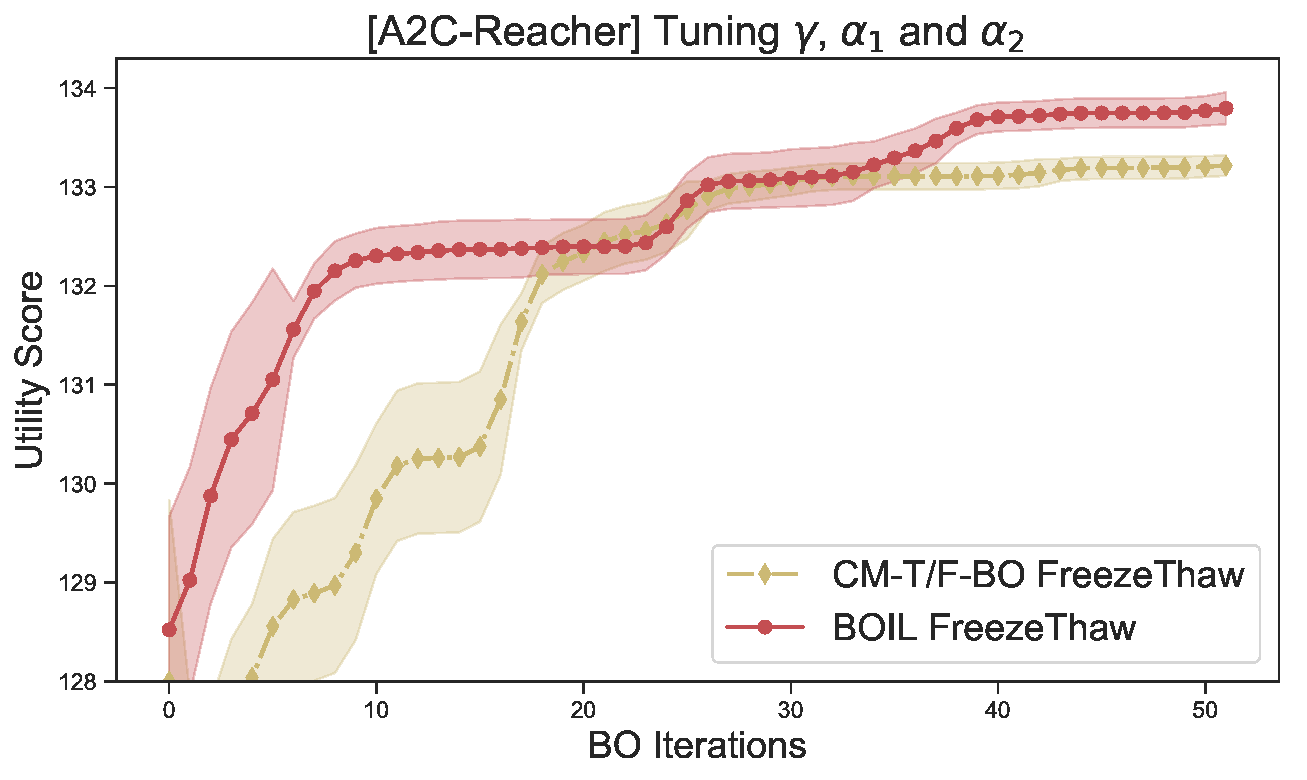}\includegraphics[width=0.5\textwidth]{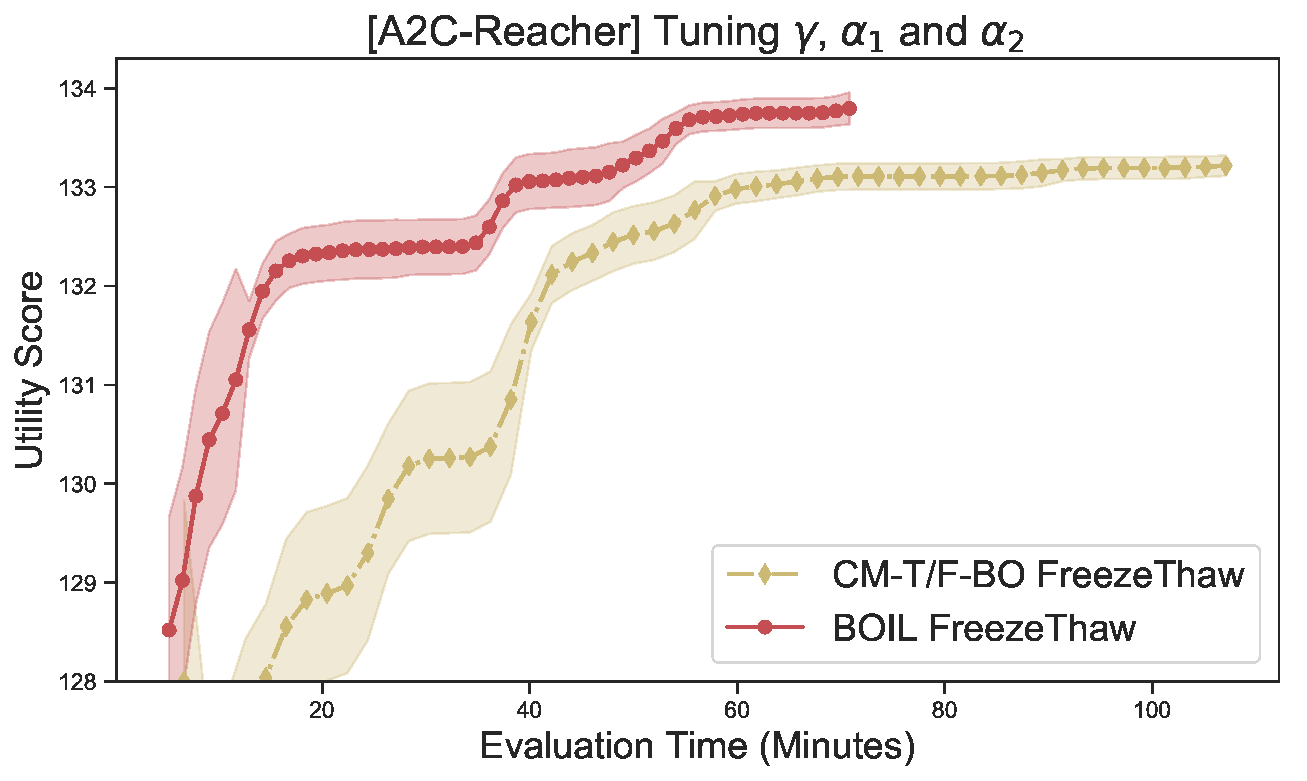}
\par\end{centering}
\begin{centering}
\includegraphics[width=0.5\textwidth]{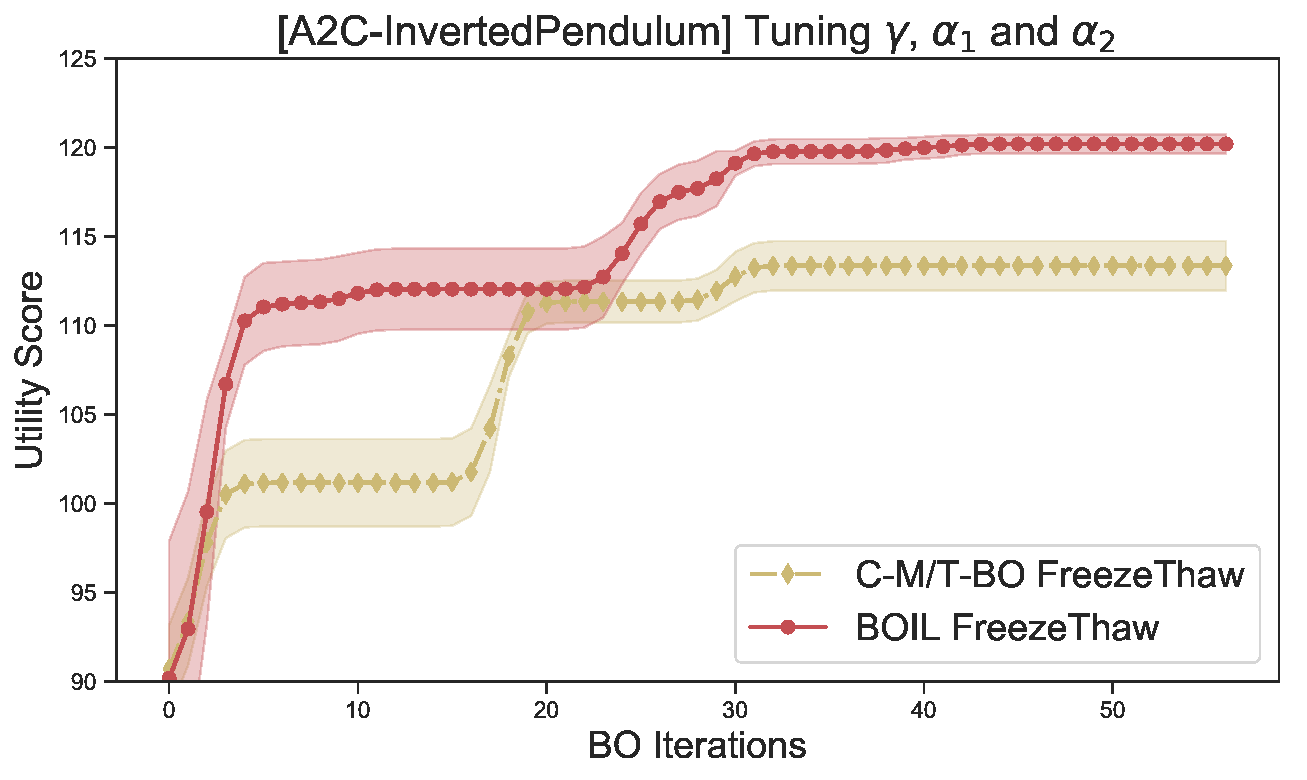}\includegraphics[width=0.5\textwidth]{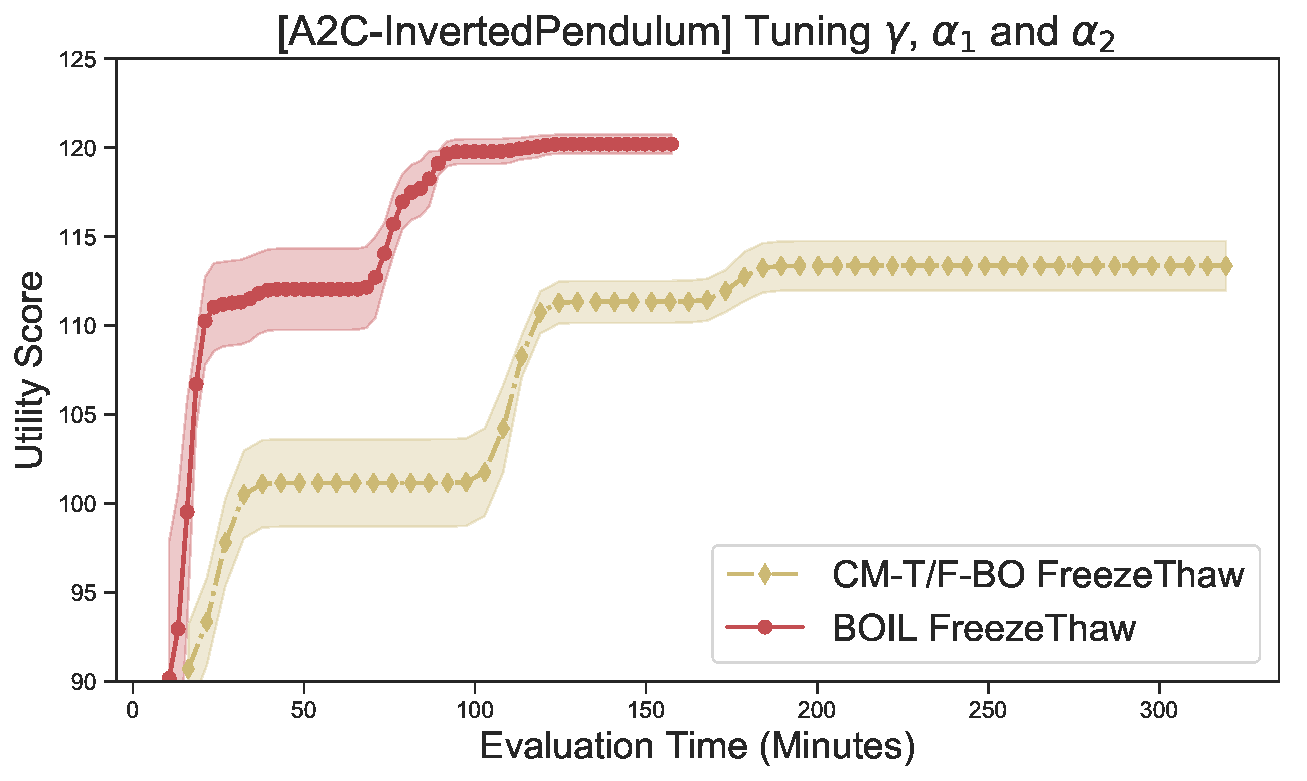}
\par\end{centering}
\caption{Comparison using freezethaw kernel for time component.\label{fig:Comparison-using-freezethaw}}
\end{figure}

We present the result in Fig. \ref{fig:Comparison-using-freezethaw}
that CM-T/F-BO is still less competitive to BOIL using this specific
time kernel.  The results again validate the robustness our approach
cross different choices of kernel.

\subsection{Additional Experiments for Tuning DRL and CNN}

We present the additional experiments for tuning a DRL model using
InvertedPendulum environment and a CNN model using a subset of CIFAR10
in Fig. \ref{fig:CIFAR10}. Again, we show that the proposed model
clearly gain advantages against the baselines in tuning hyperparameters
for model with iterative learning information available.

\subsection{Examples of Deep Reinforcement Learning Training Curves}

\begin{figure*}
\begin{centering}
\includegraphics[width=1\columnwidth]{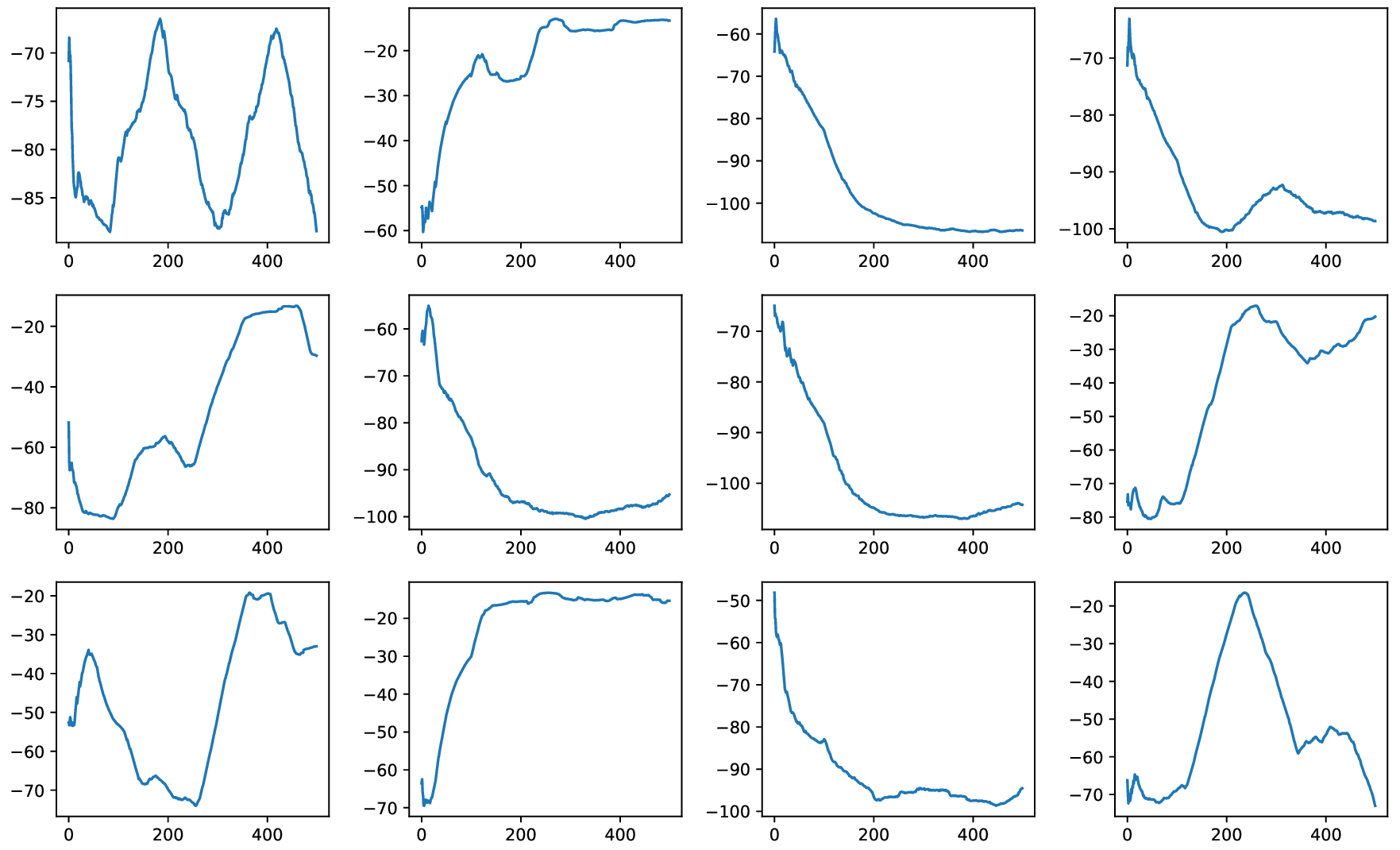}
\par\end{centering}
\begin{centering}
\includegraphics[width=1\columnwidth]{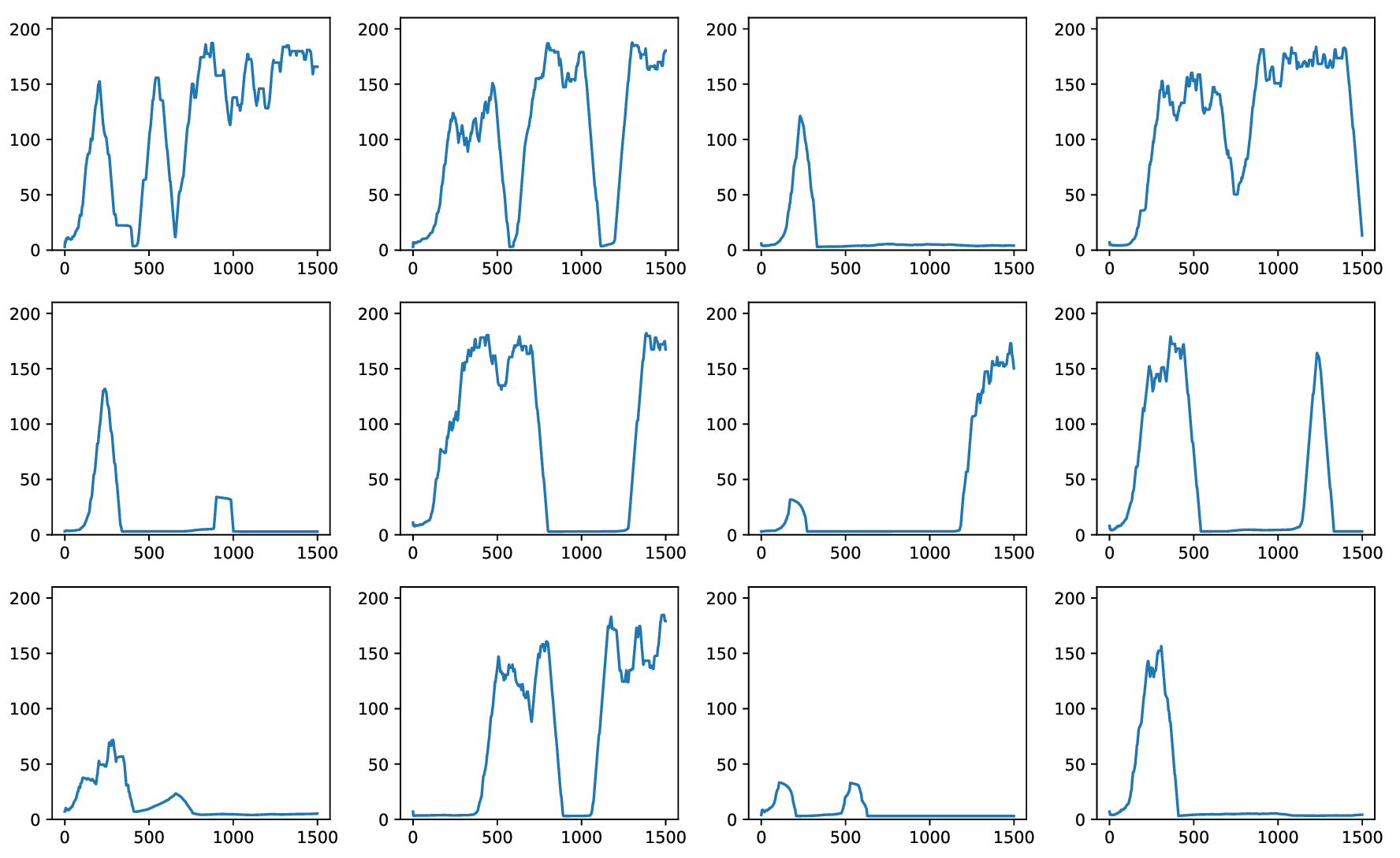}
\par\end{centering}
\caption{Examples of reward curves using A2C on Reacher-v2 (rows $1-3$) and on InvertedPendulum-v2 (rows $4-6$). Y-axis is the reward averaged over $100$ consecutive episodes. X-axis is the episode. The noisy performance illustrated is typical of DRL settings and complicates the design of early stopping criteria. Due to the property of DRL, it is not trivial to decide when to stop the training curve. In addition, it will be misleading if we only take average over the last $100$ iterations. \label{fig:Examples-of-Curves}}
\end{figure*}

Finally, we present examples of training curves produced by the deep reinforcement learning algorithm A2C in Fig. \ref{fig:Examples-of-Curves}. These fluctuate widely and it may not be trivial to define good stopping criteria as done for other applications in previous work \cite{swersky2014freeze}.

